\documentclass[journal]{IEEEtran}
\usepackage{multirow,booktabs,color,soul,threeparttable}
\definecolor{hl}{rgb}{0.75,0.75,0.75}
\sethlcolor{hl}
\usepackage{graphicx}
\usepackage{subfigure}       
\usepackage{amsmath,amsfonts,amssymb}
\usepackage{algorithm}
\usepackage{algpseudocode}   

\usepackage{stfloats}
\usepackage{verbatim}
\usepackage{array}
\usepackage{marvosym}
\usepackage{colortbl}
\usepackage[table,xcdraw]{xcolor}

\soulregister\cite7
\soulregister\citep7
\soulregister\citet7
\soulregister\ref7
\soulregister\pageref7

\begin{document}
	\title{Diffusion-based Evolutionary Optimization for 3D Multi-Objective Molecular Generation}

	\author{
		Ruiqing Sun, Sen Yang, Dawei Feng, Ronghang Wang, Huaiyuan Song, Bo Ding, Yijie Wang* and Huaimin Wang 
		\thanks{Ruiqing Sun, Dawei Feng, Ronghang Wang, Huaiyuan Song, Bo Ding, Yijie Wang (corresponding author), and Huaimin Wang are with the College of Computer Science and Technology, National University of Defense Technology, Changsha 410073, China (e-mail: sunny0331@foxmail.com; davyfeng.c@qq.com; wangronghang@nudt.edu.cn; songhuaiyuan@nudt.edu.cn; dingbo@nudt.edu.cn; wangyijie@nudt.edu.cn; hmwang@nudt.edu.cn).}
		\thanks{Sen Yang is with the Bioinformatics Center of AMMS, Beijing 100080, China (e-mail: yangsen.nudt@hotmail.com).}
	}

	\markboth{ \today}
	{Shell \MakeLowercase{\textit{et al.}}: Bare Demo of IEEEtran.cls for IEEE Journals}



	\maketitle
	
	\begin{abstract}
De novo 3D molecular design with multiple properties and fragment-retention constraints is formulated as a constrained multi-objective optimization problem. Existing 3D diffusion models provide strong capabilities for generating novel molecules, but adapting them to new conditions and multi-objective optimization tasks remains difficult. 
We present Diffusion-based Evolutionary Molecular Optimization (DEMO), a framework that draws solely on the unconditional molecular prior of a frozen pretrained diffusion model to generate chemically valid 3D candidates, while evolutionary search handles conditional generation and constrained multi-objective optimization, without retraining or fine-tuning.
DEMO performs crossover and mutation on intermediate diffusion states through an Evolutionary-Guided Diffusion (EGD) operator, combines 2D topology and 3D geometry in Structure-Aware Environmental Selection (SAES) to preserve structural diversity, and uses three cooperating populations to coordinate structural exploration, fragment-constraint satisfaction, and property optimization. Experimental results demonstrate that EGD outperforms task-specific conditional generators in property targeting; SAES preserves structural diversity while retaining competitive hypervolume in multi-objective optimization; and DEMO improves objective convergence and generates diverse feasible molecules under fragment constraints.
These results show how evolutionary search can reuse existing diffusion models for diverse molecular design requirements.
\end{abstract}
	
	
	\section{Introduction}

Practical molecular design requires generating molecules under prescribed property conditions, balancing conflicting objectives, and sometimes retaining specified fragments. Property-conditioned generation can be cast as single-objective optimization by measuring deviation from the target values, whereas property optimization with fragment constraints forms a constrained multi-objective optimization problem (CMOP). For example, fragment-based drug discovery builds lead molecules from experimentally identified binding fragments through growing, linking, or merging while preserving key molecular recognition features \cite{murray2009rise,erlanson2016twenty}. Fragment linking, scaffold hopping, and PROTAC design pose related challenges: predefined groups must be incorporated into complete molecules while the connecting structure and overall properties are optimized \cite{imrie2020deep,bekes2022protac}. These applications call for molecular design methods that can accommodate changing property conditions, objective combinations, and structural constraints.

3D diffusion models learn molecular structure distributions and generate high-quality, novel candidates. Their generative capability, however, does not by itself provide flexible optimization or conditional generation. The scope of a conditional model is tied to the properties and combinations seen during training, and performance can degrade when several conditions must be satisfied simultaneously. Adapting a model to a new task may also require retraining or fine-tuning. Moreover, it is difficult to achieve multi-objective optimization through conditional generation, especially when constraints are imposed. Fragment retention adds another difficulty: fixing a fragment during sampling can create geometric conflicts, whereas conditioning on a fragment requires generating the unknown number of remaining atoms and determining their spatial arrangements. Sampling a large number of molecules and screening them is intuitive, but the procedure is inefficient when the desired molecules must satisfy multiple conditions. Pareto-optimal molecules in the learned distribution also tend to occupy low-probability regions of the generative model. 
We therefore ask whether it is possible to solely exploit the unconditional molecular prior of existing 3D diffusion models to generate valid molecules, while delegating conditional generation and constrained multi-objective optimization to an external framework.

Evolutionary algorithms offer a natural way to organize such a framework through repeated candidate generation, objective evaluation, and population selection. Three issues must be addressed when applying them to 3D molecular design. First, position-wise crossover and mutation do not readily accommodate variable-sized molecules with no fixed atom ordering; direct coordinate recombination can disrupt bonding patterns and local geometry. Second, selection mechanisms that maintain diversity mainly through objective values may retain molecules with different objective values but similar scaffolds, leading to loss of structural diversity. Standard evolutionary algorithms can incorporate structural information, but the resulting selection mechanism must work with diffusion-based generation. Third, under fragment constraints, organizing search only by objectives and constraint violations can concentrate the population around already discovered feasible scaffolds and limit exploration elsewhere. Search must therefore coordinate structural exploration, fragment-constraint satisfaction, and feasible-solution optimization.

We present Diffusion-based Evolutionary Molecular Optimization (DEMO) to meet these requirements. DEMO connects evolutionary search to a pretrained generative model through an Evolutionary-Guided Diffusion (EGD) operator, which performs crossover and mutation on intermediate diffusion states before reverse denoising produces chemically valid molecular candidates. Structure-Aware Environmental Selection (SAES) combines 2D topological and 3D geometric similarity to limit structural redundancy while accounting for objective performance. For multi-fragment constraints, three interacting populations handle structural exploration, progressive constraint satisfaction, and property optimization of feasible solutions. Together, these components support tasks ranging from property targeting to constrained multi-objective optimization while leaving the diffusion model parameters unchanged.
The main contributions are as follows:

\begin{itemize}
    \item We propose EGD, an evolutionary variation operator that leverages a frozen pretrained 3D diffusion model to support single- and multi-property targeting without retraining.

    \item We propose SAES, which incorporates 2D topological and 3D geometric similarity into environmental selection to preserve structural diversity in multi-objective molecular search, including docking tasks.

    \item We develop DEMO, a framework that employs three cooperating populations to handle multi-fragment constraints, and evaluate it on constrained multi-objective tasks using EDM and GeoLDM as diffusion backbones.
\end{itemize}

\section{Background}
	
	\begin{figure}[htbp]
		\centering
		\includegraphics[width=0.24\textwidth]{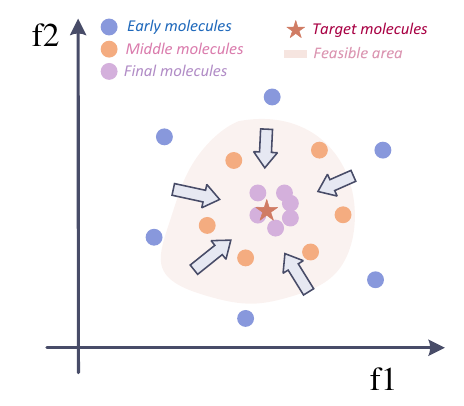}
		\includegraphics[width=0.24\textwidth]{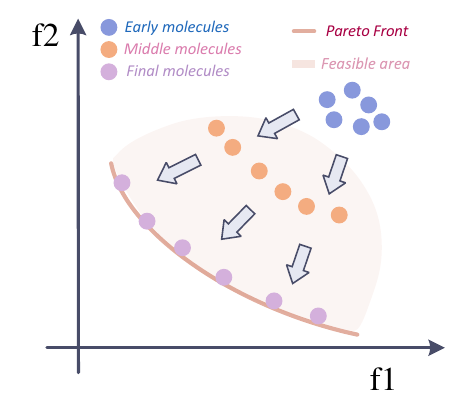}
		\caption{Illustration of single-objective optimization (SOP, left) and multi-objective optimization (MOP, right). SOP converges to one target solution. MOP progresses toward the feasible region and approximates the Pareto front.}
		\label{formulation}
	\end{figure}
	
	\subsection{3D Molecular Discovery as a CMOP}
	
	Molecular structures in 3D space are typically represented as a tuple \( M = (X, H) \), where \(X = (x_1, \dots, x_n) \in \mathbb{R}^{3 \times n}\) denotes the 3D coordinates of \(n\) atoms, and \(H = (h_1, \dots, h_n) \in \mathbb{R}^{ a \times n}\) encodes $a$-dimensional atomic features. A fundamental property of molecular systems is their invariance to translations and proper rotations of \(X\), while the coordinate generation process must be equivariant to these transformations. The combination of translations and proper rotations forms the special Euclidean group \(\mathrm{SE}(3)\), which describes rigid-body motions. Intuitively, \(\mathrm{SE}(3)\) captures all the ways one can pick up a molecule and move it around in space without changing its internal geometry. Within this group, \(\mathrm{SO}(3)\) is the special orthogonal group in three dimensions, representing pure rotations. Formally, for a rotation matrix \(R \in \mathrm{SO}(3)\) and a translation \(\hat{t} \in \mathbb{R}^3\), invariance requires \(f(RX + \hat{t}, H) = f(X, H)\), whereas equivariance requires \(g(RX + \hat{t}, H) = Rg(X, H) + \hat{t}\).
	
	Finding molecular structures with desired properties is traditionally framed as a single-objective optimization problem (SOP). This approach scalarizes multiple property goals into a single objective by minimizing a distance metric \(d\):
	\begin{equation}
		\min_{M \in \mathcal{S}} d((f_1(M), \dots, f_K(M)), \mathbf{y}^*) \quad \text{s.t.} \quad g_j(M) \le 0
	\end{equation}
	where the molecule \(M\) in the search space $\mathcal{S}$ is evaluated by \(K\) individual property predictors \(f_k(M)\), \( \mathbf{y}^*\) is the target vector, and \(g_j(M)\) represents strict structural or conformational constraints.
	
	However, real-world molecular discovery inherently requires optimizing multiple conflicting physicochemical properties, which can be formulated as a Multi-Objective Optimization Problem (MOP). A general MOP aims to optimize a set of conflicting objectives:
	\begin{equation}
		\min_{M \in \mathcal{S}} F(M) = (f_1(M), f_2(M), \dots, f_K(M))^T
	\end{equation}
	In practice, molecular design often involves additional constraints \(g_j(M)\), which are dual-natured and require both topological precision (e.g., strict structural fragment retention) and continuous property feasibility (e.g., safety or stability thresholds). When these constraints, involving discrete substructure matching and continuous property thresholds, must be explicitly satisfied during multi-objective optimization, the task is formulated as a CMOP \cite{zx1}:
	\begin{equation}
		\min_{M \in \mathcal{S}} F(M) = (f_1(M), f_2(M), \dots, f_K(M))^T \quad \text{s.t.} \quad g_j(M) \le 0
	\end{equation}
	Based on Pareto dominance, a solution \(M_1\) dominates another solution \(M_2\) (\(M_1 \succ M_2\)) if it is no worse than \(M_2\) in all objectives and better in at least one objective. The non-dominated molecules form the Pareto set, and their images under \(F\) form the Pareto front (PF). Diversity along the PF is objective-space diversity; diversity among the molecules themselves is structural diversity. In CMOPs, only feasible solutions satisfying all constraints are retained, and the objective vectors of the non-dominated feasible solutions form the constrained Pareto front (CPF) \cite{zx2}. The goal of solving a CMOP is to discover the CPF without relying on predefined scalarization weights.

	\subsection{3D Molecular Diffusion Models}
	
	3D molecular diffusion models generate molecular structures while respecting geometric symmetries. GeoDiff generates molecular conformations conditioned on molecular graphs \cite{geodiff}, whereas EDM jointly denoises atom types and Cartesian coordinates for de novo molecule generation \cite{edm}. Subsequent studies improve molecular generation through interatomic relation modeling \cite{mdm}, geometric latent diffusion \cite{geoldm}, geometry-complete and chirality-aware message passing \cite{gcdm}, and joint generation of molecular topology and geometry \cite{moldiff,mudiff}. For structure-based drug design, TargetDiff, DiffSBDD, and DecompDiff generate 3D ligands conditioned on protein pockets or structural priors \cite{targetdiff,diffsbdd,decompdiff}.
	
	Although some methods explicitly model molecular graphs and chemical bonds, most 3D geometric diffusion models represent molecules primarily through atomic features and Cartesian coordinates. The chemical graph is then inferred from the generated atom types and interatomic distances.

	Following this setting, our work adopts coordinate-based diffusion models and performs evolutionary operations on their intermediate noisy Cartesian representations.
	
	Given the molecular representation \(M=(X,H)\) defined above, the forward process gradually adds Gaussian noise over \(T\) steps according to a variance schedule \(\beta_t\):
	\begin{equation}
		q(M_t \mid M_{t-1})
		=
		\mathcal{N}\left(
		M_t;
		\sqrt{1-\beta_t}M_{t-1},
		\beta_t\mathbf{I}
		\right),
	\end{equation}
	where \(\mathbf{I}\) is the identity matrix with the same dimension as \(M_t\).
	
	The reverse process reconstructs the molecular structure from the noisy state:
	\begin{equation}
		p_\theta(M_{t-1}\mid M_t,C)
		=
		\mathcal{N}\left(
		M_{t-1};
		\mu_\theta(M_t,t,C),
		\Sigma_\theta(M_t,t,C)
		\right),
	\end{equation}
	where \(C\) denotes an optional condition. The denoising network is geometrically equivariant, ensuring that transformations of the input coordinates produce corresponding transformations of the predicted coordinates.
	
	Despite their strong generative performance, conditional diffusion models are usually trained for fixed properties or structural conditions. Their conditioning mechanisms are closely coupled with the training data and model architecture, making adaptation to unseen objectives or task-specific conditions costly. Therefore, existing diffusion models mainly serve as conditional generators rather than flexible multi-objective optimizers.

	\subsection{Evolutionary Algorithms in Molecular Discovery}
	
	Evolutionary Algorithms (EAs) are well suited to molecular discovery because they require only objective evaluations and can optimize non-differentiable or simulation-based objectives, such as molecular docking scores. Existing molecular EAs mainly operate on symbolic strings, molecular graphs, molecular fragments, or learned latent representations.
	
	Early methods apply evolutionary operators to SMILES or SELFIES strings and 2D molecular graphs. GB-GA performs chemically informed crossover and mutation directly on molecular graphs \cite{jensen2019graph}, while JANUS combines SELFIES-based molecular editing with separate exploratory and exploitative populations \cite{janus2022}. For multi-objective molecular design, graph-based Pareto optimization applies NSGA-II and NSGA-III to molecular graphs \cite{verhellen2022pareto}, whereas MOMO conducts Pareto-based evolutionary search in a learned continuous chemical space \cite{momo2024}. More recently, MOLLEO employs chemistry-aware Large Language Models (LLMs) as crossover and mutation operators \cite{molleo2025}. EAs have also been applied to structure-based drug design. AutoGrow4 evolves ligands through chemically predefined crossover and mutation operators and evaluates them using molecular docking \cite{autogrow4}. RGA further uses reinforcement learning with 3D protein--ligand representations to guide the selection of crossover parents and mutation reactions \cite{rga2022}. These operators follow predefined chemical transformations, string or graph edits, or search in a learned latent space. 
	
	Position-wise crossover and mutation remain difficult for 3D molecules. Conventional operators assume an ordered vector in which the same index is the same decision variable in every individual. A molecule is variable-sized and unordered, so atoms that share a coordinate index need not have the same structural role. Recombining raw coordinates can therefore disrupt bonding patterns and local geometry. A generated structure must also remain chemically valid and free of steric conflicts, and its coordinates are defined only up to the special Euclidean group \(\mathrm{SE}(3)\). Recall that \(\mathrm{SE}(3)\) represents rigid-body motions; therefore, the absolute Cartesian coordinates of two identical molecules can be completely different if they have different orientations or positions in space. Recombining such raw coordinates directly can severely disrupt bonding patterns and local geometry.
	
	Objective-space diversity does not by itself preserve structural diversity. Molecules with different property values can share a scaffold, so crowding measured only in objective space can fill the population with minor variants. This is not an inherent limitation of NSGA-II, NSGA-III, or SPEA2. Fingerprint distance, scaffold identity, decision-space niching, and explicit diversity archives can be added to standard environmental selection \cite{tanabe2020review,rosenthal2020diversity,verhellen2022pareto}. Multimodal multi-objective algorithms already maintain diversity in both the objective space and the decision space \cite{tanabe2020review}. Rosenthal further compares diversity-promoting selection strategies for multi-objective molecular optimization \cite{rosenthal2020diversity}. These mechanisms still need a similarity defined on molecular structure, because that diversity depends jointly on atom types, bonding topology, conformation, and spatial arrangement rather than on aligned decision variables. SAES provides this similarity by combining 2D topological and 3D geometric distances, and applies it together with diffusion-based variation. 
	
	\section{Methods}
	\begin{figure*}[htbp]
		\centering
		\includegraphics[width=0.99\textwidth]{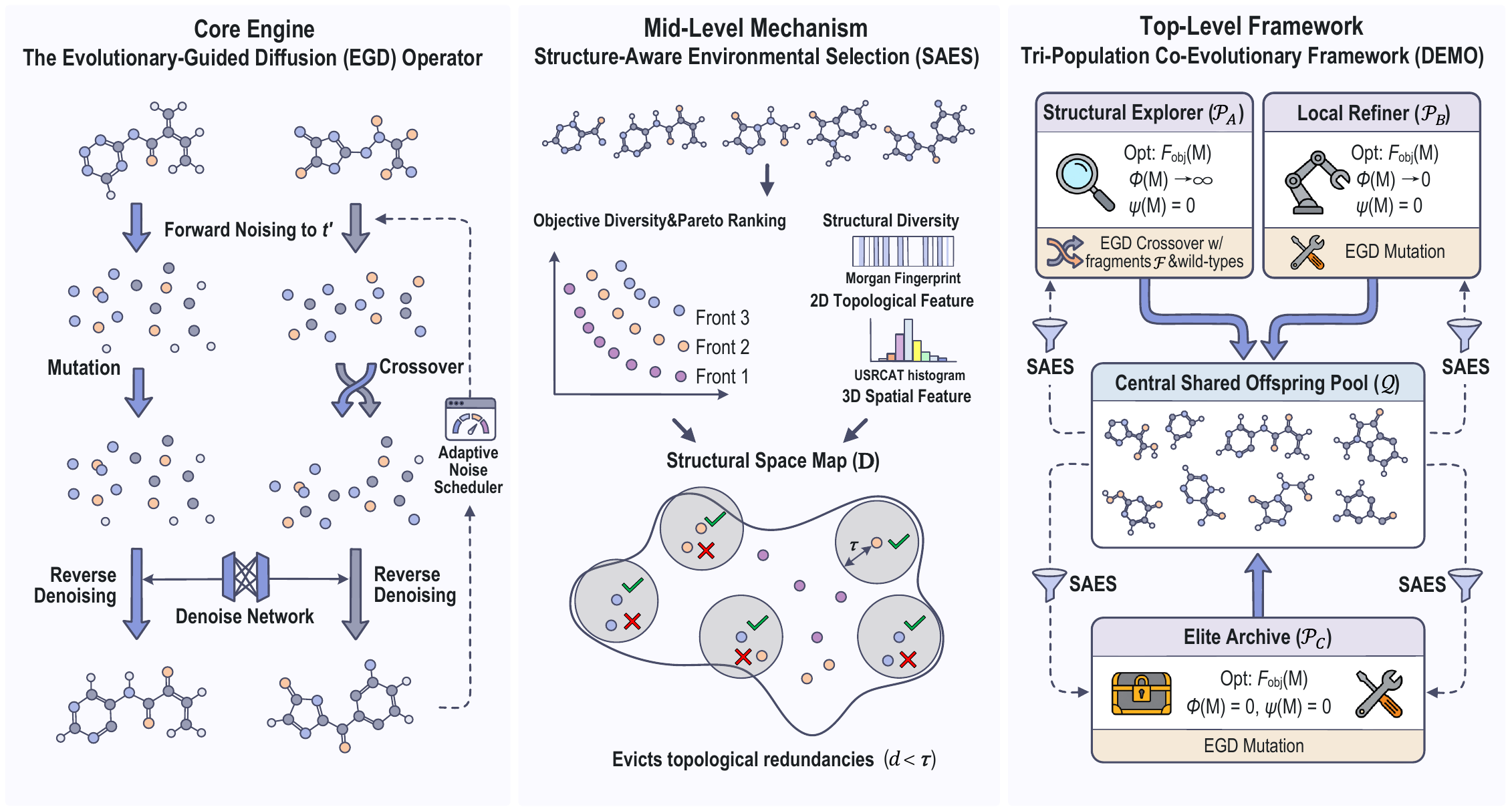}
		\caption{Overall architecture of the DEMO framework. }
		\label{fig:egd_concept}
	\end{figure*}

	\subsection{Evolutionary-Guided Diffusion Operator}
	\subsubsection{Noise-Space Crossover and Mutation}
	
	Conventional position-wise genetic operators are not directly applicable to 3D molecules because their atom sets are variable-sized and unordered. EGD instead performs crossover and mutation on intermediate noisy states of a pretrained diffusion model.
	
	Given a parent population \(\mathcal{P}\), the adaptive scheduler first selects a noise level \(t^{\prime}\). To address the \(\mathrm{SE}(3)\) ambiguity of 3D coordinates, each parent molecule is first translated to its zero-centroid coordinate frame and then diffused to the same timestep \(t^{\prime}\). EGD extracts an atomic subset from each noisy parent and concatenates them into a combined intermediate state. The resulting state is recentered again before reverse denoising to remove translational offsets introduced by subset extraction and recombination. The relative orientation is treated as part of the perturbed intermediate state. The equivariant denoiser then reconstructs the molecular geometry under the learned molecular prior without introducing dependence on the global coordinate frame.
	
	We interpret the combined state as a parent-conditioned latent prompt for reverse diffusion. Unlike an explicit condition, this prompt is introduced directly through an intermediate noisy state rather than an auxiliary conditioning network. It provides a parent-dependent intermediate state for reverse diffusion while leaving the subsequent denoising trajectory unconstrained.
	
	For mutation, the noisy state of a single parent serves as a parent-dependent intermediate state. The injected noise perturbs its atomic features and coordinates, allowing reverse diffusion to generate a new proposal from this intermediate state.
	
	Starting from either prompt $\tilde{M}_{t'}$, the pretrained network performs reverse denoising from $t'$ to $0$. The reverse process maps the perturbed state toward the learned molecular distribution and produces the offspring $\tilde{M}_0$. After all operations are completed, EGD evaluates the structural integrity of the crossover offspring and returns the score to the adaptive scheduler. Algorithm~\ref{alg:egd} summarizes the complete procedure.
	
	\begin{algorithm}[htbp]
		\caption{Evolutionary-Guided Diffusion (EGD) with Adaptive Feedback}
		\label{alg:egd}
		\begin{algorithmic}[1]
			\Require Parent population $\mathcal{P}$, forward process $q$, denoising network $p_\theta$, adaptive noise scheduler $\mathcal{S}$
			\Ensure Offspring population $\mathcal{O}$, updated scheduler $\mathcal{S}$
			\State $t' \leftarrow \mathcal{S}.\text{get\_optimal\_noise}()$, $\mathcal{O} \leftarrow \emptyset$
			\For{\textbf{each} scheduled genetic operation}
			\If{Operation == Crossover}
			\State Select parent pair $(M_0^{(1)}, M_0^{(2)})$ from $\mathcal{P}$
			\State $M_{t'}^{(i)} \sim q(\cdot \mid M_0^{(i)})$, $i \in \{1,2\}$ 
			\Comment{Forward noising}
			\State Extract random subsets $S_1 \subset M_{t'}^{(1)}$ and $S_2 \subset M_{t'}^{(2)}$
			\State $\tilde{M}_{t'} \leftarrow [S_1 \parallel S_2]$
			\Comment{Construct latent state}
			\Else
			\State Select single parent $M_0^{(1)}$ from $\mathcal{P}$
			\State $\tilde{M}_{t'} \sim q(\cdot \mid M_0^{(1)})$
			\Comment{Structural perturbation}
			\EndIf
			\For{$s = t'$ \textbf{down to} $1$}
			\State Sample $\tilde{M}_{s-1} \sim p_\theta(\cdot \mid \tilde{M}_s, s)$
			\Comment{Reverse denoising}
			\EndFor
			\State $\mathcal{O} \leftarrow \mathcal{O} \cup \{\tilde{M}_0\}$
			\EndFor
			\State $\mathcal{O}_{\mathrm{cross}} \leftarrow
			\{M \in \mathcal{O} \mid M \text{ is generated via crossover}\}$
			\State $\mathrm{Score}_{t'} \leftarrow
			\mathrm{EvaluateIntegrity}(\mathcal{O}_{\mathrm{cross}})$
			\State $\mathcal{S}.\mathrm{update\_state}
			(t', \mathrm{Score}_{t'})$
			\State \Return $\mathcal{O}, \mathcal{S}$
		\end{algorithmic}
	\end{algorithm}
	
	\subsubsection{Adaptive Noise Scheduling for EGD}
	
\begin{figure}[t]
		\centering
		\includegraphics[width=0.48\textwidth]{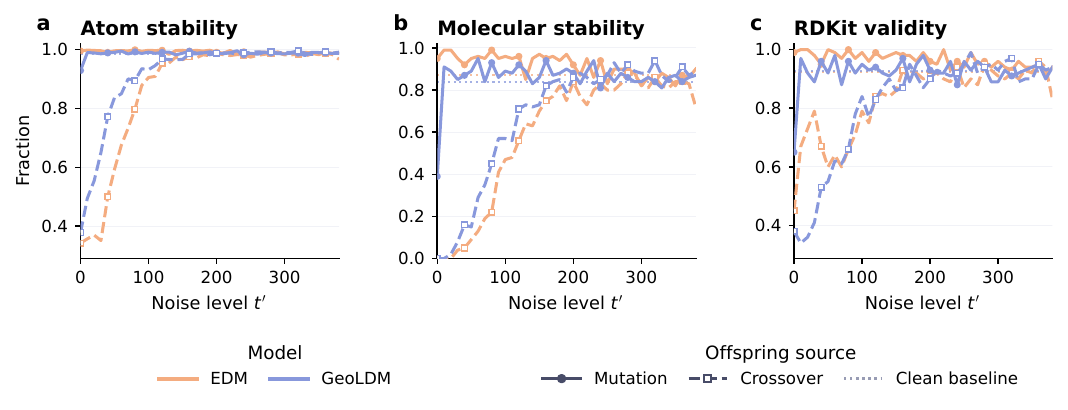}
		\caption{Effect of the noise level $t'$ on EGD offspring quality on QM9. Molecular validity and atom stability increase rapidly before reaching a plateau.}
		\label{fig:noise_quality_1}
	\end{figure}

	The noise level $t'$ controls the perturbation strength before reverse diffusion. If $t'$ is too small, the denoising process has limited freedom to repair the irregular intermediate states created by crossover. This leads to low molecular validity and atom stability. Increasing $t'$ improves structural recovery, but excessive noise produces unnecessarily stochastic proposals and approaches unconditional generation. As shown in Figure~\ref{fig:noise_quality_1}, generation quality increases rapidly with $t'$ and then reaches a plateau. We define its left boundary as the validity knee point.
	
	\begin{figure}[htbp]
		\centering
		\includegraphics[width=0.48\textwidth]{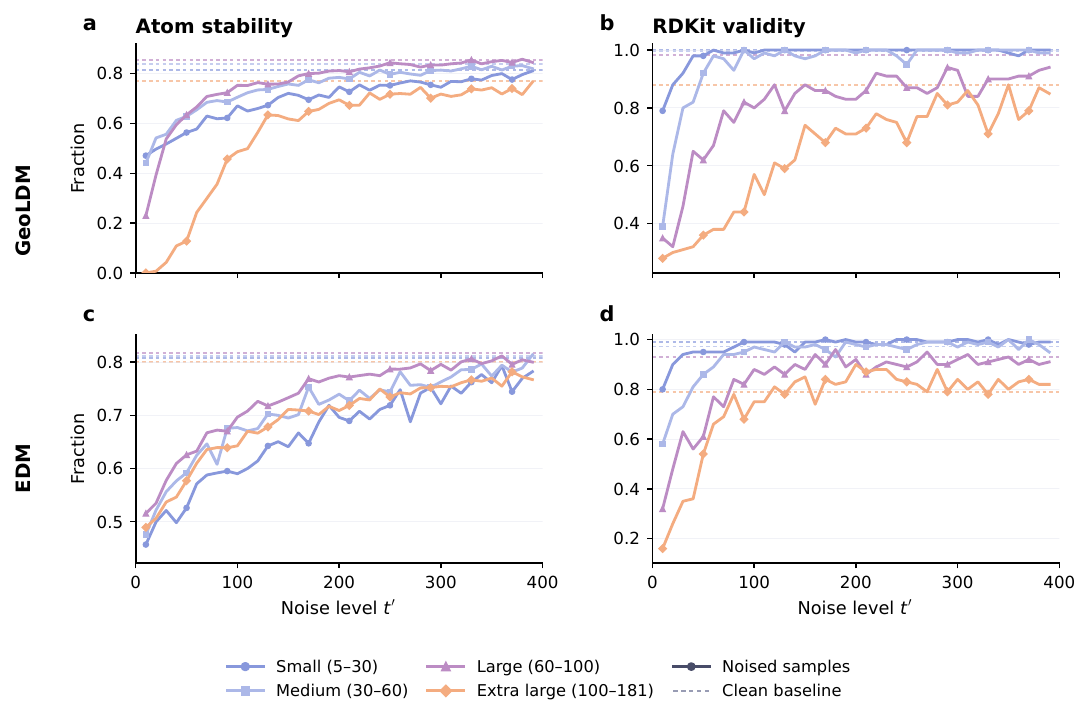}
		\caption{Effect of molecular size $N$ on noise robustness (GEOM-Drugs). The validity knee point shifts toward larger $t'$ as molecular size increases, with consistent trends for GeoLDM and EDM.}
		\label{fig:n_dependency}
	\end{figure}
	
	The appropriate $t'$ also depends on molecular size. As shown in Figure~\ref{fig:n_dependency}, molecules with 5--30 atoms recover structural integrity at approximately $t'=150$, whereas molecules with 100--181 atoms require noise levels above $250$. The same trend appears for both GeoLDM and EDM. A fixed noise level is therefore unsuitable for molecules with different sizes and complexities.
	
	We propose an Adaptive Noise Scheduler to locate the validity knee point during evolution. The scheduler evaluates crossover offspring using molecular validity and atom stability, which measure molecule-level and atom-level chemical integrity, respectively. These two metrics jointly reflect the generation quality of 3D molecular diffusion models. We therefore define
	\[
	\mathrm{Score}(t')=\mathrm{ValidityRate}(t')\times\mathrm{AtomStabilityRate}(t').
	\]
	The multiplicative form assigns a high score only when both criteria are satisfied. This score essentially evaluates how the noise level $t'$ affects the model's ability to map the perturbed state back to the learned molecular distribution. The scheduler starts from $t_{\mathrm{curr}}=T/2$ in the \textsc{Binary\_Search} mode. It first searches toward lower noise levels to locate the validity transition and then switches to Gaussian Process (GP) optimization for fine adjustment. The complete procedure is given in Algorithm~\ref{alg:scheduler}.
	
	\begin{algorithm}[htbp]
		\caption{Adaptive Noise Scheduler}
		\label{alg:scheduler}
		\begin{algorithmic}[1]
			\Require Current noise level $t_{\mathrm{curr}}$, current score $\mathrm{Score}_{\mathrm{curr}}$, persistent state $\{\mathrm{Mode},t_{\mathrm{low}},t_{\mathrm{high}},\mathrm{SafeBaseline},\mathcal{D}\}$, noise bounds $[t_{\min},t_{\max}]$, GP model with a Matérn kernel, and parameters $\mathrm{DropFactor}$, $\beta$, $\lambda$, and $\sigma_{\mathrm{jitter}}$
			\Ensure Noise level $t_{\mathrm{next}}$ and updated scheduler state
			\State $\mathcal{H} \leftarrow \{s \mid (t,s)\in\mathcal{D}\}$
			\State $\mathcal{D} \leftarrow \mathcal{D}\cup\{(t_{\mathrm{curr}},\mathrm{Score}_{\mathrm{curr}})\}$
			\If{$\mathrm{Mode} == \textsc{Binary\_Search}$}
			\If{$\mathcal{H}\neq\emptyset$ \textbf{and} $\mathrm{Score}_{\mathrm{curr}} < \mathrm{DropFactor}\times\mathrm{Mean}(\mathcal{H})$}
			\State $t_{\mathrm{low}} \leftarrow t_{\mathrm{curr}}$
			\State $\mathrm{Mode} \leftarrow \textsc{Adaptive\_GP}$
			\State $t_{\mathrm{next}} \leftarrow \mathrm{SafeBaseline}$
			\Else
			\State $\mathrm{SafeBaseline} \leftarrow t_{\mathrm{curr}}$
			\State $t_{\mathrm{high}} \leftarrow t_{\mathrm{curr}}$
			\State $t_{\mathrm{next}} \leftarrow (t_{\mathrm{low}}+t_{\mathrm{high}})/2$
			\EndIf
			\ElsIf{$\mathrm{Mode} == \textsc{Adaptive\_GP}$}
			\State Fit the GP model on $\mathcal{D}$
			\State Define $t_{\mathrm{cand}}\in[t_{\min},t_{\max}]$
			\State Predict $\mu(t_{\mathrm{cand}})$ and $\sigma(t_{\mathrm{cand}})$
			\State $a(t) \leftarrow \mu(t)+\beta\sigma(t)-\lambda\left(\frac{t-t_{\min}}{t_{\max}-t_{\min}}\right)$
			\State $t^* \leftarrow \arg\max_{t_{\mathrm{cand}}}a(t_{\mathrm{cand}})$
			\State $t_{\mathrm{next}}\sim\mathcal{N}(t^*,\sigma_{\mathrm{jitter}})$
			\State Clip $t_{\mathrm{next}}$ to $[t_{\min},t_{\max}]$
			\EndIf
			\State \Return $t_{\mathrm{next}}$ and updated scheduler state
		\end{algorithmic}
	\end{algorithm}

	At the beginning of evolution, the scheduler is initialized once with $\mathrm{Mode}=\textsc{Binary\_Search}$, $t_{\mathrm{curr}}=T/2$, $t_{\mathrm{low}}=t_{\min}$, $t_{\mathrm{high}}=T/2$, $\mathrm{SafeBaseline}=T/2$, and $\mathcal{D}=\emptyset$. These variables are maintained as persistent scheduler states and updated once per generation.
	The scheduler starts from $t'=T/2$ and progressively tests lower noise levels. As long as offspring quality remains stable, the current value is recorded as the latest safe baseline and the search continues toward smaller $t'$. A sharp score decrease relative to previous observations indicates that the search has crossed the validity knee point. The scheduler then returns to the last safe baseline and switches to GP-based adjustment.
	
	The GP uses a Matérn kernel with $\nu=1.5$ to model the potentially sharp transition in $\mathrm{Score}(t')$. Its acquisition function combines the Upper Confidence Bound with a linear noise penalty. The UCB term balances predicted generation quality and model uncertainty, while the penalty discourages unnecessarily large noise levels. Gaussian jitter prevents repeated selection of the same $t'$. The scheduler therefore tracks the left side of the validity plateau and maintains high-quality offspring without introducing excessive perturbation.

	\subsubsection{Discussion}
	
	\textbf{Why avoid step-wise structural constraints?}
	EGD crossover resembles RePaint in that both initialize reverse diffusion from a partially specified state. RePaint is well suited to images because the missing pixels and the fixed pixels occupy known positions on a regular grid. This assumption does not hold for molecular generation: the number, types, and relative positions of the missing atoms are unknown, and the observed fragment has no fixed absolute pose under SE(3) transformations. Replacing the fragment at every denoising step forcibly distorts the denoising trajectory, leading to geometrically implausible molecular fragments. In our single-fragment experiment, this strategy caused severe geometric conflicts: both EDM and GeoLDM obtained zero molecular stability, with atom stability below $0.69$ and validity below $0.46$. EGD instead inserts the latent prompt only once at timestep $t'$ and leaves the subsequent reverse process unconstrained, allowing the network to jointly adjust the complete structure and recover molecular validity.
	
	\textbf{Why use the same noise scheduling for crossover and mutation?}
	Although mutation starts from one valid parent, insufficient noise often causes reverse diffusion to reconstruct a nearly identical molecule because the parent remains within the same local attraction region of the learned distribution. This produces limited variation and weakens evolutionary exploration. Mutation therefore uses the same scheduler-selected noise level $t'$ as crossover to introduce sufficient perturbation and generate distinct candidates. The adaptive scheduler adjusts this level according to offspring integrity, allowing both operators to maintain molecular validity while providing effective search variation.

	\subsection{Structure-Aware Environmental Selection (SAES)}
	
	\begin{algorithm}[htpb]
		\caption{Structure-Aware Environmental Selection (SAES)}
		\label{alg:saes}
		\begin{algorithmic}[1]
			\Require Candidate pool $\mathcal{Q}$, target size $K$, objective vector $\mathbf{z}(M)$, threshold $\tau$
			\Ensure Selected population $\mathcal{P}$
			\State Compute structural distance matrix $\mathbf{D}$ for $\mathcal{Q}$ using Eq.~(\ref{eq:structural_distance})
			\State Compute SPEA2 fitness array $Fit$ from $\mathbf{z}(M)$ and $\mathbf{D}$
			\State $\mathcal{P} \leftarrow \emptyset,\quad \mathcal{U} \leftarrow \mathcal{Q}$
			\State $\mathcal{B} \leftarrow \{i\in\mathcal{U}\mid Fit_i=\min_{u\in\mathcal{U}}Fit_u\}$
			\State $i^* \leftarrow \arg\max_{i\in\mathcal{B}}\left(\frac{1}{|\mathcal{U}|-1}\sum_{\substack{j\in\mathcal{U}\\j\neq i}}\mathbf{D}_{i,j}\right)$
			\State Move $M_{i^*}$ from $\mathcal{U}$ to $\mathcal{P}$
			\While{$|\mathcal{P}|<K$}
			\State $d_j \leftarrow \min_{M_k\in\mathcal{P}}\mathbf{D}_{j,k},\quad \forall M_j\in\mathcal{U}$
			\State $\mathcal{V} \leftarrow \{j\in\mathcal{U}\mid d_j\ge\tau\}$
			\If{$\mathcal{V}\neq\emptyset$}
			\State $f^* \leftarrow \min_{v\in\mathcal{V}}Fit_v$
			\State $j^* \leftarrow \arg\max_{\{j\in\mathcal{V}\mid Fit_j=f^*\}}d_j$
			\Else
			\State $f^* \leftarrow \min_{u\in\mathcal{U}}Fit_u$
			\State $j^* \leftarrow \arg\max_{\{j\in\mathcal{U}\mid Fit_j=f^*\}}d_j$
			\EndIf
			\State Move $M_{j^*}$ from $\mathcal{U}$ to $\mathcal{P}$
			\EndWhile
			\State \Return $\mathcal{P}$
		\end{algorithmic}
	\end{algorithm}
	
	After generating molecular candidates with EGD, environmental selection must balance optimization performance and structural diversity. Conventional EMO algorithms mainly estimate population density in the objective space, which does not necessarily reflect molecular structural diversity. We therefore incorporate molecular structure directly into environmental selection.
	
	SAES combines 2D topology and 3D geometry. For each molecule, we compute a 2048-bit radius-2 Morgan fingerprint and a 60-dimensional USRCAT descriptor. The structural distance between molecules $M_a$ and $M_b$ is defined as
	\begin{equation}
		\label{eq:structural_distance}
		D(M_a,M_b)
		=
		\frac{1}{2}D_{\mathrm{Jaccard}}
		(\mathrm{FP}_a,\mathrm{FP}_b)
		+
		\frac{1}{2}\widehat{D}_{\mathrm{Euc}}
		(\mathrm{USR}_a,\mathrm{USR}_b),
	\end{equation}
	where $D_{\mathrm{Jaccard}} = 1 - |\mathrm{FP}_a\cap\mathrm{FP}_b|\,/\,|\mathrm{FP}_a\cup\mathrm{FP}_b|$ is the Jaccard distance between the two fingerprints, computed from the sets of activated fingerprint bits (0 for identical fingerprints and close to 1 when no bit is shared), and $\widehat{D}_{\mathrm{Euc}}$ denotes the Euclidean-distance matrix normalized by its maximum value within the current candidate pool.
	
	SAES adopts an SPEA2-style fitness $Fit_i=R_i+1/(d_{i,k}+2)$, where $R_i$ is the raw fitness computed from Pareto dominance in the population-specific objective space $\mathbf{z}(M)$, and $d_{i,k}$ is the distance to the $k$-th nearest neighbor in the molecular structural space, with $k=\max(1,\lfloor\sqrt{|\mathcal{Q}|}\rfloor)$. A smaller $Fit_i$ indicates better fitness.
	
	Algorithm~\ref{alg:saes} summarizes the selection procedure. The initial seed is selected from the candidates with minimum fitness, with ties resolved by the largest mean structural distance to the candidate pool. Subsequent candidates must satisfy a minimum distance threshold $\tau$ from the selected population. Among eligible candidates, fitness is prioritized and structural distance is used as a tie-breaker. If no candidate satisfies $\tau$, the threshold constraint is waived for that iteration.

	\subsection{The Diffusion-based Evolutionary Molecular Optimization Framework (DEMO)}
	
	\begin{algorithm}[htbp]
		\caption{Tri-Population DEMO}
		\label{alg:demo}
		\begin{algorithmic}[1]
			\Require Diffusion model $p_\theta$, target fragments $\mathcal{F}$, base objectives $F(M)$, population size $N$
			\Ensure Feasible Pareto archive $\mathcal{P}_C$
			\State $\mathcal{P}_0 \leftarrow \{M\in\mathrm{Evaluate}(\mathrm{Sample}(N))\mid\psi(M)=0\}$
			\State $\mathcal{P}_A,\mathcal{P}_B,\mathcal{P}_C \leftarrow \mathrm{InitializePopulations}(\mathcal{P}_0)$
			\For{$g=1,\ldots,G$}
			\State $t' \leftarrow \mathcal{S}.\mathrm{GetNoiseLevel}()$
			\State $\mathcal{O}_A \leftarrow \mathrm{EGD_{Crossover}}(\mathcal{P}_A,\mathcal{F},t',N/2)$
			\State $\mathcal{O}_{BC} \leftarrow \mathrm{EGD_{Mutation}}(\mathcal{P}_B,\mathcal{P}_C,t',N/2)$
			\State $(\mathcal{E}_A,\mathcal{E}_{BC}) \leftarrow \mathrm{DenoiseAndEvaluate}(\mathcal{O}_A,\mathcal{O}_{BC},p_\theta)$
			\State $\mathcal{Q} \leftarrow \{M\in\mathcal{P}_A\cup\mathcal{P}_B\cup\mathcal{E}_A\cup\mathcal{E}_{BC}\mid\psi(M)=0\}$
			\State $\mathbf{z}_A(M) \leftarrow [F(M),-\phi(M)]$
			\State $\mathcal{P}_A \leftarrow \mathrm{SAES}(\mathcal{Q},N,\mathbf{z}_A,\tau)$
			\State $\mathcal{Q}_B \leftarrow \{M\in\mathcal{Q}\mid\phi(M)>0\}$
			\State $\mathbf{z}_B(M) \leftarrow [F(M),\phi(M)]$
			\State $\mathcal{P}_B \leftarrow \mathrm{SAES}(\mathcal{Q}_B,N,\mathbf{z}_B,\tau)$
			\State $\mathcal{Q}_C \leftarrow \{M\in\mathcal{Q}\cup\mathcal{P}_C\mid\phi(M)=0\}$
			\State $\mathcal{P}_C \leftarrow \mathrm{SAES}(\mathcal{Q}_C,N,F,\tau)$
			\State $\mathcal{S}.\mathrm{Update}(t',\mathrm{Integrity}(\mathcal{E}_A))$
			\EndFor
			\State \Return $\mathcal{P}_C$
		\end{algorithmic}
	\end{algorithm}
	
	While SAES promotes structural diversity, assembling multiple target fragments requires coordinated exploration and refinement. DEMO therefore maintains three interacting populations: the Structural Explorer ($\mathcal{P}_A$), the Local Refiner ($\mathcal{P}_B$), and the Elite Archive ($\mathcal{P}_C$). Each population uses SAES with objectives and constraints corresponding to its evolutionary role.
	
	Let $\mathcal{F}$ denote the target fragments, $\phi(M)$ the structural penalty measuring incomplete fragment incorporation, $\psi(M)$ the global chemical-validity constraint, and $F(M)$ the original optimization objectives. Molecules with $\psi(M)>0$ are discarded before environmental selection. The Structural Explorer uses $\mathrm{Objs}=[F(M),-\phi(M)]$ to explore structurally diverse regions away from the target fragments. The Local Refiner considers molecules with $\phi(M)>0$ and uses $\mathrm{Objs}=[F(M),\phi(M)]$ to progressively approach structural feasibility. The Elite Archive contains only fully feasible molecules satisfying $\phi(M)=0$ and $\psi(M)=0$ and optimizes $F(M)$ alone.
	
	For offspring generation, selected parents from $\mathcal{P}_A$ are divided between internal crossover and fragment-conditioned crossover. The former combines representations sampled from $\mathcal{P}_A$, whereas the latter combines parent and target-fragment representations before reverse denoising. In contrast, $\mathcal{P}_B$ and $\mathcal{P}_C$ dynamically share a fixed mutation budget. Mutation in $\mathcal{P}_B$ refines partially assembled candidates, while mutation in $\mathcal{P}_C$ further optimizes the objectives of feasible solutions.
	
	All crossover and mutation intermediates are processed together by the SE(3)-equivariant reverse diffusion model. The evaluated offspring are combined with the resident populations to form a central pool. Population-specific filtering, fitness evaluation, and SAES are then applied independently to update $\mathcal{P}_A$, $\mathcal{P}_B$, and $\mathcal{P}_C$. Algorithm~\ref{alg:demo} summarizes the complete procedure.

\section{Experiments}

\textbf{Tasks and Datasets.}
We evaluate DEMO on five types of tasks: Single-Property Targeting,
Multi-Property Targeting, Unconstrained Multi-Objective Optimization
(MOP), fragment-constrained Multi-Objective Optimization (CMOP), and
3D protein--ligand docking. EDM \cite{edm} and GeoLDM
\cite{geoldm} are used as the diffusion backbones. They are pretrained
on QM9 \cite{qm9} for the property-targeting, MOP, and CMOP tasks and
on GEOM-Drugs \cite{drugs} for docking. To prevent data leakage, QM9
is equally divided into two disjoint subsets for training the
diffusion models and property predictors, respectively. For docking,
Vina score, QED, and SA are jointly optimized across 10 diverse
protein pockets selected from CrossDocked2020 \cite{cd2020}.

For the multi-fragment CMOP, we construct 10 property combinations
for each diffusion backbone. Since the target fragment pairs are
selected separately for GeoLDM and EDM, this evaluation contains 20
independent tasks. Results are therefore compared only within the
same backbone.

\textbf{Evaluation Metrics.}
Property-targeting performance is evaluated using Mean Absolute Error
(MAE). For MOP and CMOP tasks, Hypervolume (HV) measures Pareto-front
convergence. Structural diversity is evaluated using the Unique Valid
Rate (UVR) for MOPs and the Unique Feasible Rate (UFR) for CMOPs. UVR
is the proportion of valid molecules with unique molecular structures,
whereas UFR is the proportion of structurally unique molecules that
satisfy all constraints, identified by canonical SMILES. The canonical SMILES retain stereochemical information, so stereoisomers are counted as distinct molecules, whereas different conformers of the same molecule are counted as duplicates. UVR and UFR measure non-duplication among the returned molecules rather than the magnitude of structural diversity, which is evaluated separately using internal Tanimoto diversity, scaffold counts, and 3D motif diversity (Figure~\ref{fig:mop_diversity_metrics}). For scale-independent HV calculation, objectives are min--max normalized using dataset-derived lower and upper bounds. The normalization bounds are obtained from QM9 for property-based tasks. Values outside the dataset range are not clipped and may therefore fall outside $[0,1]$. For docking, the bounds are $[0,1]$ for SA and QED and $[\mathrm{native\_ligand\_score},0]$ for the Vina score. The reference point is $\mathbf{r}=[1,\ldots,1]^T$.

\textbf{Baselines and Computational Budgets.}
The baselines include four categories:
(1) \textit{Conditional Generation Models}: cEDM, cGeoLDM, cGCDM, NExT-Mol \cite{nextmol}, PropMolFlow \cite{propmolflow} and EEGSDE \cite{eegsde};
(2) \textit{Gradient-Guided Methods}: TFG \cite{tfg} and MUDM \cite{mudm};
(3) \textit{Generate-and-Screen Methods}: Top-$N$ draws molecules from the unconditional diffusion model and retains the $N$ candidates with the best task objective values. It does not apply crossover, mutation, or iterative selection. The Same Evaluations (SE) budget matches the number of property-predictor calls used by the evolutionary run, and the Same Runtime (SR) budget matches the number of denoising steps.
(4) \textit{Constrained Multi-Objective Evolutionary Algorithms}:
SPEA2-CDP \cite{spea2}, CCMO \cite{ccmo}, and CMOEA-CD
\cite{cmoeacd}. To isolate the effect of SAES, all conventional
evolutionary frameworks use EGD as their molecular variation operator.

\textbf{Implementation Details.}
All evolutionary methods use a population size of $N=32$. Single-property targeting, multi-property targeting, and unconstrained MOP use 10, 20, and 25 generations, respectively. CMOP and docking each use 50 generations. SAES uses a structural-distance threshold of
$\tau=0.05$. All results are averaged over 20 independent runs. For MOP, docking, and CMOP tasks, statistical significance is assessed using the Wilcoxon rank-sum test ($p<0.05$). Symbols ($+$, $-$, $=$) indicate significantly better, worse, and statistically equivalent HV relative to the corresponding full method.

\textbf{Constraint Evaluation.}
Each CMOP task requires a generated molecule to contain two target
fragments, each consisting of seven atoms. The same fragment
constraints are used by all competing methods within each task.
Because the number of linker atoms and the relative arrangement of
the fragments are unknown before generation, structural feasibility
is evaluated through subgraph matching rather than fixed 3D
coordinates. Therefore, the fragment constraints in this work specify the retention of molecular substructures, while their conformations and relative spatial arrangements are free to adapt during 3D molecular generation.

Let $\mathcal{F}=\{F_1,\ldots,F_n\}$ denote the required target
fragments. The structural constraint violation of a generated molecule
$M$ is defined as
\begin{equation}
	CV_{\phi}(M)
	=
	1-
	\frac{
		\sum_{i=1}^{|\mathcal{F}|}
		|\operatorname{match}(M,F_i)|
	}{
		\sum_{i=1}^{|\mathcal{F}|}|F_i|
	},
	\label{eq:fragment_cv}
\end{equation}
where $|\operatorname{match}(M,F_i)|$ is the number of atoms in
$F_i$ matched within $M$. A molecule is structurally feasible if and
only if $CV_{\phi}(M)=0$. Target fragments are allowed to overlap
within the generated molecule.
We use a two-stage procedure to reduce the cost of repeated subgraph
matching. RDKit's \texttt{FindMCS} first estimates the match
percentage. If the percentage exceeds $80\%$, exact VF2 subgraph
isomorphism is performed; otherwise, half of the preliminary match
percentage is returned as a penalized matching score.

In addition to the structural constraints defined above for CMOPs,
chemical feasibility is evaluated for generated molecules in all
tasks. Every task applies constraint dominance: a reasonable molecule always outranks an unreasonable one, and unreasonable molecules are filtered out before objective comparison. For all QM9-based tasks, a molecule is reasonable only after successful RDKit sanitization, molecular stability, and atom
stability. Atom stability is the fraction of atoms with a chemically valid valence. Molecule stability is the fraction of molecules in which every atom is stable. These scores reflect the model's ability to generate chemically reasonable solutions. Docking uses a stricter version of the same filter. A molecule must also receive a negative Vina
score and pass all 20 PoseBusters \cite{posebuster} clash and
valency checks.
	
	\subsection{Results on Single- and Multi-Property Targeting}

	Single- and multi-property targeting minimize the deviation from a prescribed target on QM9:
	\begin{equation}
		\min_{M\in\mathcal{S}}|f(M)-y^*|, \quad \text{and} \quad \min_{M\in\mathcal{S}}\|\mathbf{f}(M)-\mathbf{y}^*\|,
		\label{eq:targeting}
	\end{equation}
	where \(f(M)\) is a single property (e.g., polarizability \(\alpha\), HOMO--LUMO gap \(\Delta\varepsilon\), orbital energy \(\varepsilon_{\mathrm{homo}}\) or \(\varepsilon_{\mathrm{lumo}}\), dipole moment \(\mu\), or heat capacity \(C_v\)), \(\mathbf{f}(M)\) is a property pair, and \(y^*\) and \(\mathbf{y}^*\) are the corresponding scalar and vector targets. The pairs are those listed in Table~\ref{tab:multi_property}, and Tables~\ref{tab:single_property} and~\ref{tab:multi_property} report the mean absolute error of each property.

	\begin{figure*}[b]
		\centering
		\subfigure[Top-N (SE)]{\includegraphics[width=0.18\textwidth]{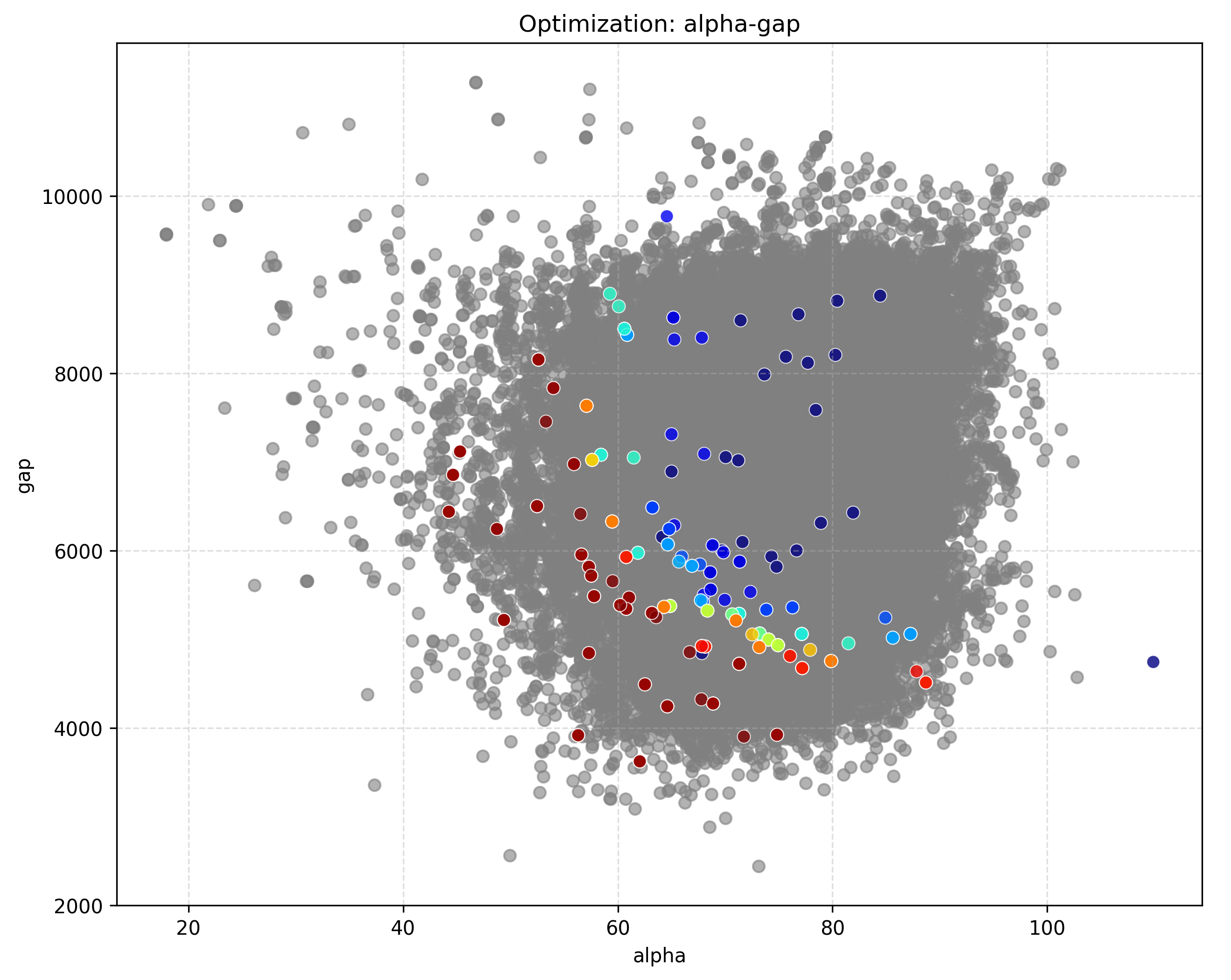}}
		\hfill
		\subfigure[SPEA2]{\includegraphics[width=0.18\textwidth]{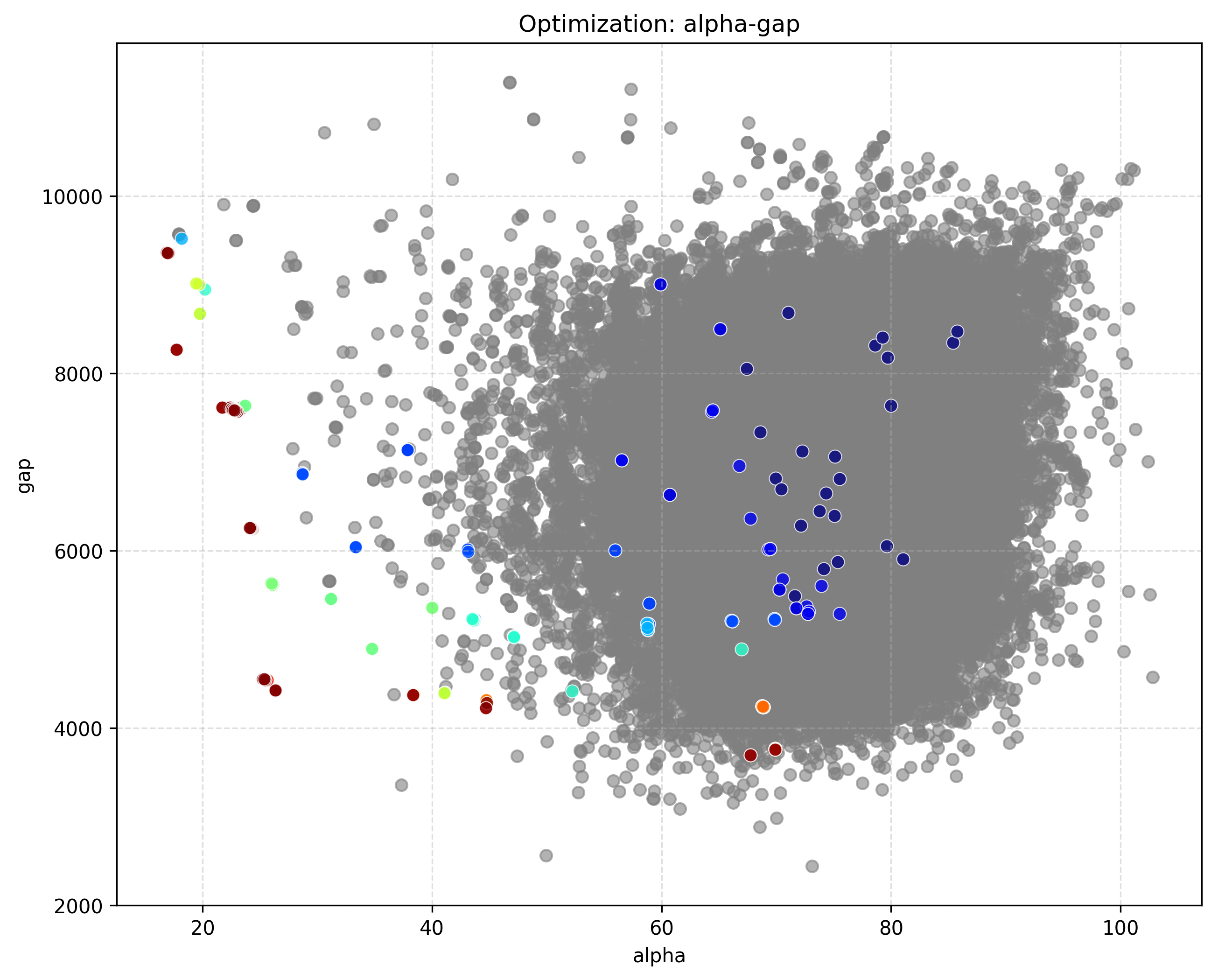}}
		\hfill
		\subfigure[SAES w/o MT]{\includegraphics[width=0.18\textwidth]{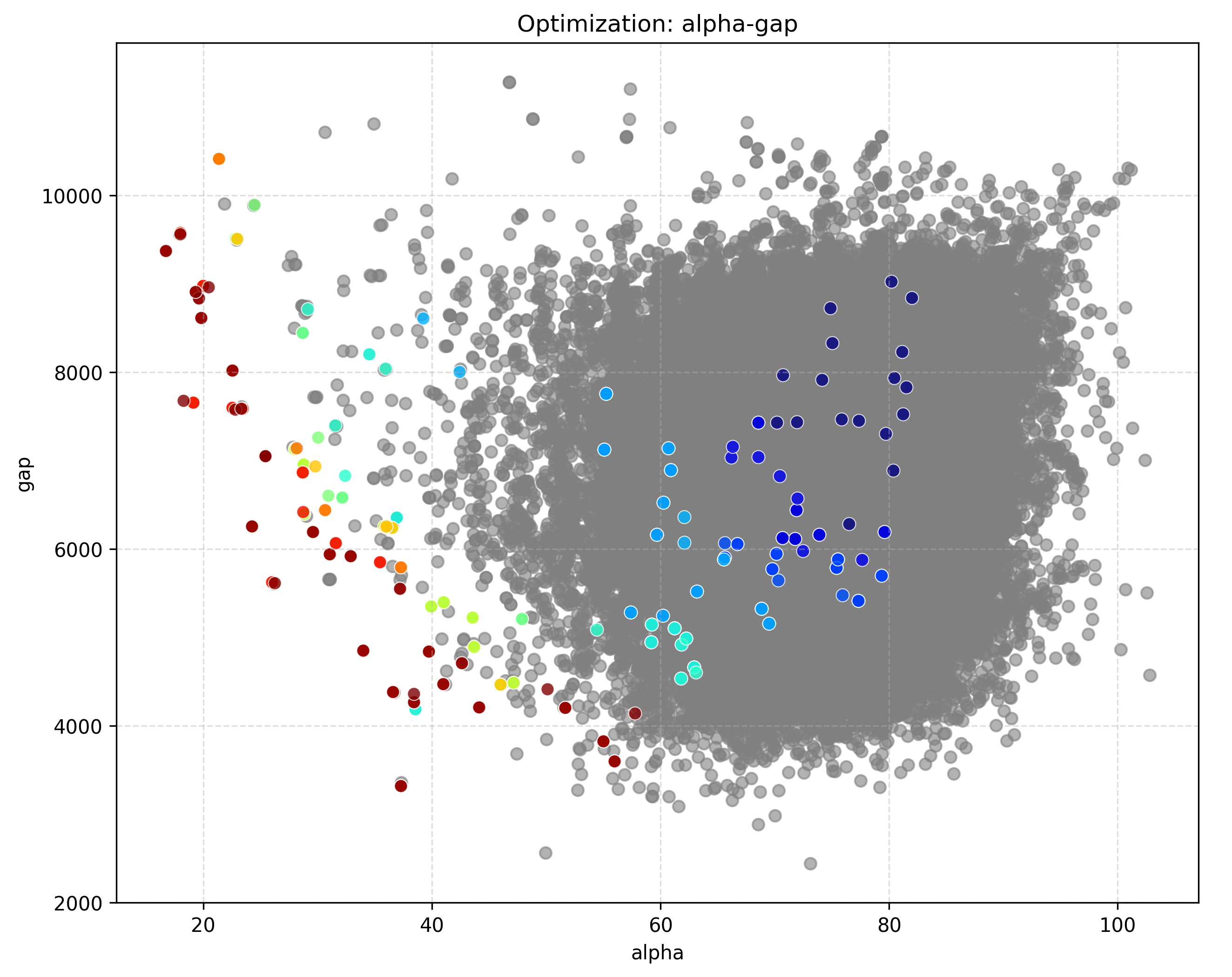}}
		\hfill
		\subfigure[SAES w/o CO]{\includegraphics[width=0.18\textwidth]{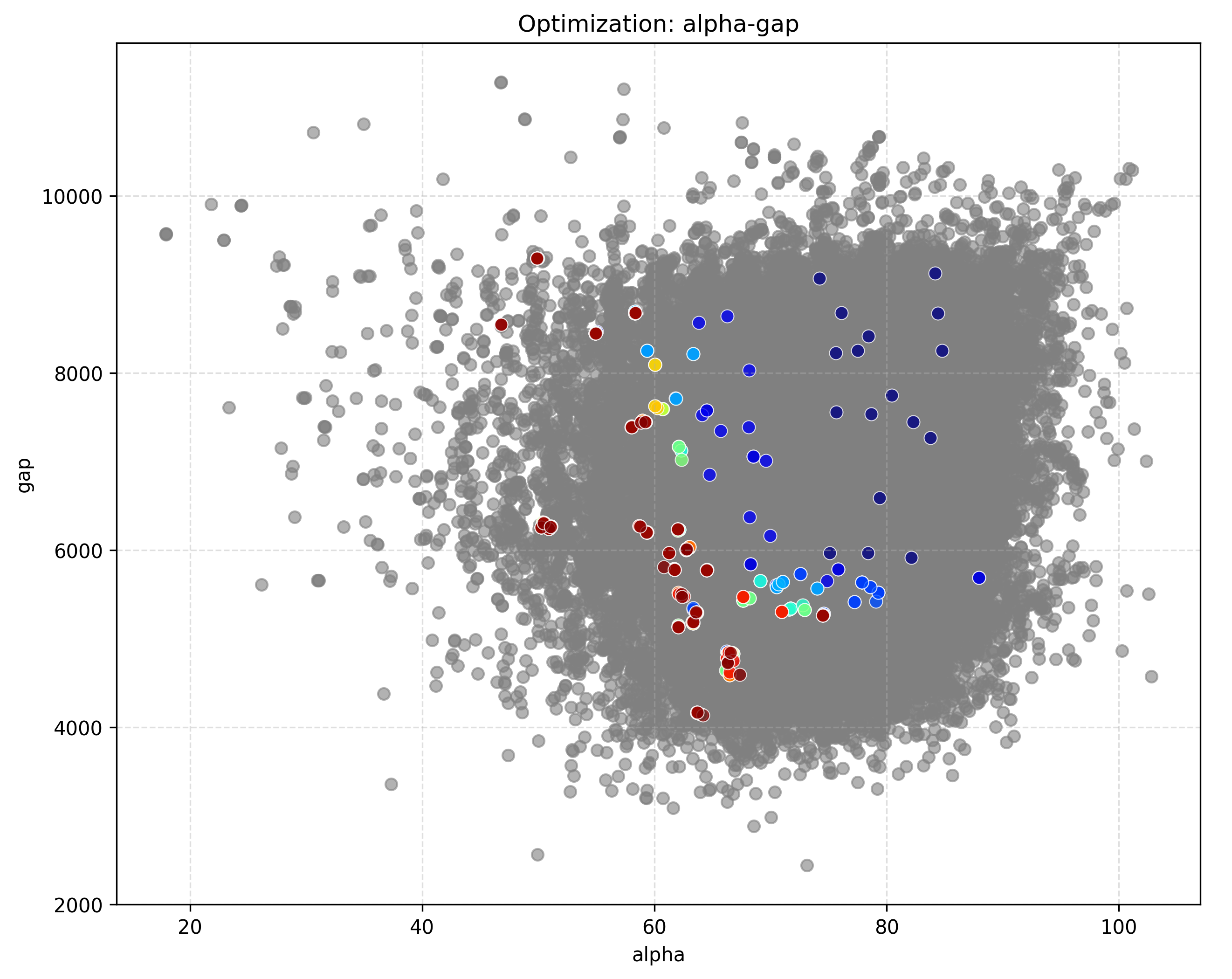}}
		\hfill
		\subfigure[SAES (Ours)]{\includegraphics[width=0.21\textwidth]{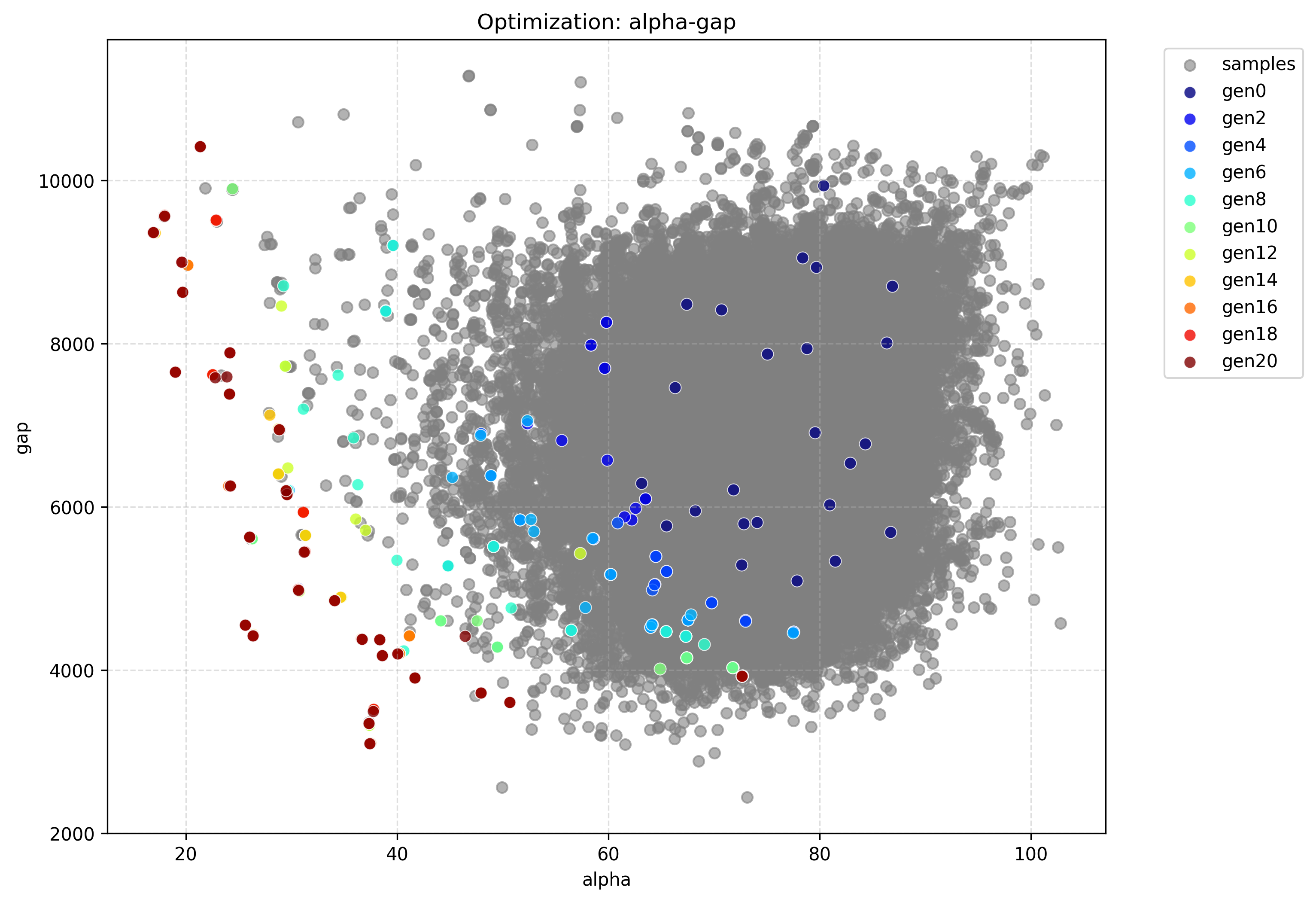}}
		\caption{Search trajectories for simultaneous optimization of $\alpha$ and $\Delta\varepsilon$. Gray points denote 100,000 molecules sampled from the unconditional model. Top-N remains concentrated in the high-density region, whereas the evolutionary methods progressively approach the Pareto boundary. Compared with SPEA2, SAES maintains a more structurally diverse set of solutions along the front.}
		\label{fig:MOP_distribution}
	\end{figure*}

	\begin{table}[b]
		\tiny
		\centering
		\setlength{\aboverulesep}{0pt}
		\setlength{\belowrulesep}{0pt}
		\caption{MAE ($\downarrow$) on Single-Property Targeting tasks.}
		\label{tab:single_property}
		\resizebox{\columnwidth}{!}{
			\begin{tabular}{lcccccc}
				\toprule
				\textbf{Method} & $\alpha$ & $\Delta\varepsilon$ & $\varepsilon_{\text{homo}}$ & $\varepsilon_{\text{lumo}}$ & $\mu$ & $C_v$ \\
				\midrule
				cEDM              & 2.76 & 655  & 356 & 584  & 1.11 & 1.10 \\
				cGeoLDM           & 2.37 & 587  & 340 & 522  & 1.11 & 1.03 \\
				EEGSDE            & 2.50 & 487  & 302 & 447  & 0.78 & 0.94 \\
				cGCDM             & 1.99 & 595  & 346 & 480  & 0.86 & 0.70 \\
				PropMolFlow       & 1.36 & 391  & 246 & 312  & 0.63 & 0.64 \\
				NExT-Mol          & 1.16 & 297  & 205 & 235  & 0.51 & 0.51 \\
				\midrule
				TFG               & 3.90 & 893  & 984 & 568  & 1.33 & 2.77 \\
				MUDM              & \underline{0.43} & 85   & 72  & 133  & 0.33 & 0.29 \\
				\midrule
				TFG+TopN (SR)     & 2.81 & 550  & 322 & 407  & 0.74 & 1.84 \\
				TFG+TopN (SE)     & 0.52 & 132  & 148 & 87   & 0.25 & 0.34 \\
				EDM+TopN (SR)     & 2.48 & 437  & 201 & 359  & 0.55 & 1.16 \\
				EDM+TopN (SE)     & 0.89 & 174  & 97  & 129  & 0.26 & \textbf{0.10} \\
				GeoLDM+TopN (SR)  & 2.23 & 390  & 212 & 314  & 0.54 & 1.57 \\
				GeoLDM+TopN (SE)  & 0.85 & 144  & 99  & 120  & 0.22 & 0.65 \\
				\midrule
				EGD+GeoLDM (Ours) & 0.88 & \underline{33} & \textbf{15} & \underline{30} & \textbf{0.03} & \underline{0.13} \\
				EGD+EDM (Ours)    & \textbf{0.14} & \textbf{31} & \underline{24} & \textbf{26} & \underline{0.05} & \textbf{0.10} \\
				\bottomrule
			\end{tabular}
		}
	\end{table}

	\begin{table}[htbp]
		\centering
		\caption{MAE ($\downarrow$) on the Multi-Property Targeting task.}
		\label{tab:multi_property}
		\resizebox{\columnwidth}{!}{
			\begin{tabular}{lc ccc cc ccc}
				\toprule
				\multirow{2}{*}{\textbf{Task}} & \multirow{2}{*}{\textbf{Prop.}} & \multicolumn{3}{c}{\textbf{Baselines}} & \multicolumn{2}{c}{\textbf{Top-N}} & \multicolumn{3}{c}{\textbf{EGD \& Ablation}} \\
				\cmidrule(lr){3-5} \cmidrule(lr){6-7} \cmidrule(lr){8-10}
				& & MUDM & EEGSDE & Base & (SE) & (SR) & w/o CO & w/o MT & \textbf{Ours} \\
				\midrule
				\multicolumn{10}{c}{\textit{Backbone: GeoLDM}} \\
				\midrule
				\multirow{2}{*}{$C_v / \mu$} 
				& $C_v$ & 1.47 & 0.98 & 1.23 & 1.21 & 2.25 & \underline{0.64} & 1.67 & \textbf{0.69} \\
				& $\mu$ & 0.69 & 0.91 & 1.12 & 0.34 & 0.59 & \underline{0.28} & 0.29 & \textbf{0.09} \\
				\midrule
				\multirow{2}{*}{$\Delta\varepsilon / \mu$} 
				& $\Delta\varepsilon$ & 544 & 563 & 664 & 234 & 461 & \textbf{78} & 228 & \underline{186} \\
				& $\mu$ & 0.58 & 0.87 & 1.13 & 0.31 & 0.54 & \textbf{0.25} & \underline{0.27} & 0.48 \\
				\midrule
				\multirow{2}{*}{$\alpha / \mu$} 
				& $\alpha$ & 1.32 & 2.61 & 2.77 & 4.14 & 2.15 & \textbf{0.86} & 2.98 & \underline{0.96} \\
				& $\mu$ & 0.52 & 0.86 & 1.09 & 0.73 & 0.40 & \underline{0.22} & 0.33 & \textbf{0.12} \\
				\midrule
				\multirow{2}{*}{$\varepsilon_{\text{ho}} / \varepsilon_{\text{lu}}$} 
				& $\varepsilon_{\text{ho}}$ & 317 & 355 & 384 & \underline{115} & 214 & \textbf{69} & 143 & 182 \\
				& $\varepsilon_{\text{lu}}$ & 455 & 517 & 634 & 238 & 412 & \textbf{143} & 211 & \underline{205} \\
				\midrule
				\multirow{2}{*}{$\varepsilon_{\text{lu}} / \mu$} 
				& $\varepsilon_{\text{lu}}$ & 575 & 526 & 636 & 318 & 509 & \underline{92} & 274 & \textbf{90} \\
				& $\mu$ & 0.50 & 0.86 & 1.06 & 0.36 & 0.60 & \underline{0.18} & 0.28 & \textbf{0.06} \\
				\midrule
				\multirow{2}{*}{$\varepsilon_{\text{lu}} / \Delta\varepsilon$} 
				& $\varepsilon_{\text{lu}}$ & 361 & 546 & 457 & 261 & 443 & \underline{155} & 239 & \textbf{88} \\
				& $\Delta\varepsilon$ & 228 & 589 & 548 & 241 & 433 & \underline{136} & 240 & \textbf{54} \\
				\midrule
				\multirow{2}{*}{$\varepsilon_{\text{ho}} / \Delta\varepsilon$} 
				& $\varepsilon_{\text{ho}}$ & 262 & 567 & 361 & 126 & 230 & \textbf{54} & 175 & \underline{62} \\
				& $\Delta\varepsilon$ & 489 & 323 & 657 & 203 & 402 & \underline{170} & 205 & \textbf{92} \\
				
				\midrule
				\multicolumn{10}{c}{\textit{Backbone: EDM}} \\
				\midrule
				\multirow{2}{*}{$C_v / \mu$} 
				& $C_v$ & 1.47 & 0.98 & 1.08 & 1.32 & 2.21 & \textbf{0.45} & 1.83 & \underline{0.57} \\
				& $\mu$ & 0.69 & 0.91 & 1.16 & 0.35 & 0.62 & \underline{0.31} & 0.82 & \textbf{0.07} \\
				\midrule
				\multirow{2}{*}{$\Delta\varepsilon / \mu$} 
				& $\Delta\varepsilon$ & 544 & 563 & 683 & 247 & 581 & \underline{106} & 415 & \textbf{101} \\
				& $\mu$ & 0.58 & 0.87 & 1.16 & 0.30 & 0.63 & \underline{0.18} & 0.86 & \textbf{0.09} \\
				\midrule
				\multirow{2}{*}{$\alpha / \mu$} 
				& $\alpha$ & \underline{1.32} & 2.61 & 2.76 & 2.07 & 4.32 & \textbf{0.94} & 2.29 & 1.63 \\
				& $\mu$ & 0.52 & 0.86 & 1.16 & 0.40 & 0.69 & \underline{0.25} & \textbf{0.21} & 0.66 \\
				\midrule
				\multirow{2}{*}{$\varepsilon_{\text{ho}} / \varepsilon_{\text{lu}}$} 
				& $\varepsilon_{\text{ho}}$ & 317 & 355 & 372 & \underline{122} & 277 & \underline{68} & 131 & \textbf{44} \\
				& $\varepsilon_{\text{lu}}$ & 455 & 517 & 594 & 231 & 421 & \underline{148} & 238 & \textbf{91} \\
				\midrule
				\multirow{2}{*}{$\varepsilon_{\text{lu}} / \mu$} 
				& $\varepsilon_{\text{lu}}$ & 575 & 526 & 610 & 298 & 505 & \underline{78} & 252 & \textbf{58} \\
				& $\mu$ & 0.50 & 0.86 & 1.14 & 0.34 & 0.61 & \underline{0.20} & 0.35 & \textbf{0.06} \\
				\midrule
				\multirow{2}{*}{$\varepsilon_{\text{lu}} / \Delta\varepsilon$} 
				& $\varepsilon_{\text{lu}}$ & 361 & 546 & 1097& 269 & 463 & \underline{93} & 212 & \textbf{86} \\
				& $\Delta\varepsilon$ & 228 & 589 & 712 & 252 & 415 & \textbf{137} & \underline{172} & 178 \\
				\midrule
				\multirow{2}{*}{$\varepsilon_{\text{ho}} / \Delta\varepsilon$} 
				& $\varepsilon_{\text{ho}}$ & 262 & 567 & 578 & 125 & 237 & \underline{122} & 184 & \textbf{48} \\
				& $\Delta\varepsilon$ & 489 & 323 & 655 & 240 & 450 & \underline{117} & 250 & \textbf{43} \\
				\bottomrule
			\end{tabular}
		}
	\end{table}
	
	Tables~\ref{tab:single_property} and~\ref{tab:multi_property} evaluate the property-search capability of EGD. Across the two diffusion backbones, EGD achieves the best or tied-best result on all six properties. It substantially outperforms task-specific conditional models. EGD uses only pretrained unconditional diffusion models and requires no retraining or fine-tuning. It also generally outperforms the train-free guidance methods TFG and MUDM.
	
	The limitation of train-free-guidance-based methods becomes more evident in multi-property targeting. Methods such as MUDM need predefined guidance weights derived from the property statistics of the training data when multiple targets are optimized jointly. These weights are used to balance different guidance signals. Without appropriate prior weights, the guidance directions can conflict and performance can degrade substantially \cite{mudm}. EGD does not require such prior knowledge. It directly evaluates generated molecules and uses evolutionary selection to coordinate multiple target properties.

	Top-$N$ sampling provides another contrast. It relies entirely on candidates drawn from the unconditional distribution. It therefore has difficulty reaching narrow target regions, especially when the desired values lie in low-density intersections of multiple properties. Even under the larger SE budget, Top-$N$ generally produces higher errors than EGD. EGD instead repeatedly selects promising candidates and applies diffusion-based variation. This gradually redirects the population toward the desired property region.
	
	The ablation results in Table~\ref{tab:multi_property} further clarify the roles of crossover and mutation. Removing mutation generally weakens targeting performance. This indicates that mutation is important for refining candidates near the desired property vector. The mutation-only variant, EGD w/o CO, occasionally outperforms the full method. Multi-property targeting requires precise convergence to a specified point. Local perturbations can therefore provide more stable refinement. In contrast, crossover generates higher-variance proposals and may temporarily move candidates away from the target. Its advantage becomes more evident when broader exploration is required, as shown in the subsequent multi-objective optimization experiments.

	\subsection{Results on MOPs}

	\begin{table*}[hbpt]
		\centering
		\caption{HV ($\uparrow$) and UVR ($\uparrow$) on unconstrained multi-objective optimization.
			Values in square brackets denote UVR.
			Symbols ($+$, $-$, $=$) indicate the statistical significance of HV relative to the full SAES method. $HV_D$ and $HV_{100K}$ denote the HV of the reference dataset and 100,000 unconditional samples, respectively.}
		\label{tab:mop}
		
		\setlength{\tabcolsep}{2.5pt}
		\renewcommand{\arraystretch}{1.08}
		
		\resizebox{\textwidth}{!}{
			\begin{tabular}{l cc cc ccccc cc ccccc}
				\toprule
				
				\multirow{3}{*}{\textbf{Task}}
				& \multirow{3}{*}{\textbf{HV$_D$}}
				& \multirow{3}{*}{\textbf{HV$_{100K}$}}
				& \multicolumn{7}{c}{\textbf{GeoLDM}}
				& \multicolumn{7}{c}{\textbf{EDM}} \\
				
				\cmidrule(lr){4-10}
				\cmidrule(lr){11-17}
				
				& & 
				& \multicolumn{2}{c}{\textbf{Top-N}}
				& \multicolumn{5}{c}{\textbf{EGD Ablation \& Selection}}
				& \multicolumn{2}{c}{\textbf{Top-N}}
				& \multicolumn{5}{c}{\textbf{EGD Ablation \& Selection}} \\
				
				\cmidrule(lr){4-5}
				\cmidrule(lr){6-10}
				\cmidrule(lr){11-12}
				\cmidrule(lr){13-17}
				
				& & 
				& (SR) & (SE)
				& w/o CO & w/o MT & SPEA2 & SPEA2-SD & \textbf{SAES}
				& (SR) & (SE)
				& w/o CO & w/o MT & SPEA2 & SPEA2-SD & \textbf{SAES} \\
				
				\midrule
				
				\multirow{2}{*}{$\alpha$--$\Delta\varepsilon$}
				& \multirow{2}{*}{0.854}
				& \multirow{2}{*}{0.802}
				& 0.559\tiny{(0.017)}\textsuperscript{$-$}
				& 0.590\tiny{(0.024)}\textsuperscript{$-$}
				& 0.568\tiny{(0.054)}\textsuperscript{$-$}
				& \underline{0.760}\tiny{(0.015)}\textsuperscript{$=$}
				& 0.758\tiny{(0.019)}\textsuperscript{$=$}
				& 0.759\tiny{(0.017)}
				& \textbf{0.764}\tiny{(0.016)}
				& 0.535\tiny{(0.016)}\textsuperscript{$-$}
				& 0.629\tiny{(0.029)}\textsuperscript{$-$}
				& 0.545\tiny{(0.042)}\textsuperscript{$-$}
				& \underline{0.761}\tiny{(0.012)}\textsuperscript{$=$}
				& 0.766\tiny{(0.012)}\textsuperscript{$-$}
				& 0.769\tiny{(0.016)}
				& \textbf{0.772}\tiny{(0.019)} \\
				
				& & 
				& [0.997] & [1.000] & [0.734] & [0.895] & [0.312] & [0.278] & [0.896]
				& [1.000] & [1.000] & [0.698] & [0.932] & [0.300] & [0.277] & [0.880] \\
				
				\multirow{2}{*}{$\Delta\varepsilon$--$\varepsilon_{\mathrm{ho}}$}
				& \multirow{2}{*}{0.702}
				& \multirow{2}{*}{0.678}
				& 0.461\tiny{(0.020)}\textsuperscript{$-$}
				& 0.543\tiny{(0.015)}\textsuperscript{$-$}
				& 0.489\tiny{(0.042)}\textsuperscript{$-$}
				& \textbf{0.644}\tiny{(0.019)}\textsuperscript{$=$}
				& 0.635\tiny{(0.018)}\textsuperscript{$-$}
				& 0.637\tiny{(0.027)}
				& \underline{0.639}\tiny{(0.018)}
				& 0.475\tiny{(0.020)}\textsuperscript{$-$}
				& 0.536\tiny{(0.026)}\textsuperscript{$-$}
				& 0.478\tiny{(0.036)}\textsuperscript{$-$}
				& \textbf{0.641}\tiny{(0.020)}\textsuperscript{$=$}
				& 0.640\tiny{(0.017)}\textsuperscript{$=$}
				& 0.648\tiny{(0.024)}
				& 0.640\tiny{(0.017)} \\
				
				& &
				& [0.982] & [1.000] & [0.776] & [0.962] & [0.251] & [0.247] & [0.876]
				& [1.000] & [1.000] & [0.725] & [0.951] & [0.276] & [0.231] & [0.871] \\
				
				\multirow{2}{*}{$\varepsilon_{\mathrm{ho}}$--$\varepsilon_{\mathrm{lu}}$}
				& \multirow{2}{*}{0.799}
				& \multirow{2}{*}{0.769}
				& 0.485\tiny{(0.042)}\textsuperscript{$-$}
				& 0.537\tiny{(0.025)}\textsuperscript{$-$}
				& 0.563\tiny{(0.075)}\textsuperscript{$-$}
				& \textbf{0.772}\tiny{(0.026)}\textsuperscript{$+$}
				& 0.751\tiny{(0.031)}\textsuperscript{$=$}
				& 0.737\tiny{(0.033)}
				& \underline{0.753}\tiny{(0.034)}
				& 0.470\tiny{(0.048)}\textsuperscript{$-$}
				& 0.550\tiny{(0.032)}\textsuperscript{$-$}
				& 0.525\tiny{(0.057)}\textsuperscript{$-$}
				& \underline{0.766}\tiny{(0.027)}\textsuperscript{$=$}
				& 0.740\tiny{(0.029)}\textsuperscript{$-$}
				& 0.758\tiny{(0.039)}
				& \textbf{0.767}\tiny{(0.034)} \\
				
				& &
				& [0.992] & [1.000] & [0.731] & [0.917] & [0.126] & [0.091] & [0.868]
				& [1.000] & [1.000] & [0.754] & [0.946] & [0.096] & [0.127] & [0.884] \\
				
				\multirow{2}{*}{$\varepsilon_{\mathrm{lu}}$--$\mu$}
				& \multirow{2}{*}{0.998}
				& \multirow{2}{*}{0.992}
				& 0.811\tiny{(0.032)}\textsuperscript{$-$}
				& 0.861\tiny{(0.021)}\textsuperscript{$-$}
				& 0.845\tiny{(0.046)}\textsuperscript{$-$}
				& \textbf{0.962}\tiny{(0.033)}\textsuperscript{$=$}
				& 0.948\tiny{(0.032)}\textsuperscript{$-$}
				& 0.945\tiny{(0.048)}
				& \underline{0.957}\tiny{(0.035)}
				& 0.800\tiny{(0.022)}\textsuperscript{$-$}
				& 0.858\tiny{(0.051)}\textsuperscript{$-$}
				& 0.813\tiny{(0.052)}\textsuperscript{$-$}
				& \textbf{0.983}\tiny{(0.037)}\textsuperscript{$+$}
				& \underline{0.957}\tiny{(0.040)}\textsuperscript{$=$}
				& 0.947\tiny{(0.039)}
				& 0.956\tiny{(0.039)} \\
				
				& &
				& [1.000] & [1.000] & [0.737] & [0.978] & [0.218] & [0.225] & [0.929]
				& [0.999] & [0.998] & [0.754] & [0.973] & [0.203] & [0.186] & [0.906] \\
				
				\multirow{2}{*}{$\mu$--$C_v$}
				& \multirow{2}{*}{1.000}
				& \multirow{2}{*}{0.659}
				& 0.848\tiny{(0.027)}\textsuperscript{$-$}
				& 0.886\tiny{(0.019)}\textsuperscript{$-$}
				& 0.831\tiny{(0.046)}\textsuperscript{$-$}
				& \textbf{0.992}\tiny{(0.024)}\textsuperscript{$+$}
				& 0.983\tiny{(0.016)}\textsuperscript{$=$}
				& 0.983\tiny{(0.021)}
				& \underline{0.984}\tiny{(0.016)}
				& 0.845\tiny{(0.040)}\textsuperscript{$-$}
				& 0.874\tiny{(0.031)}\textsuperscript{$-$}
				& 0.836\tiny{(0.038)}\textsuperscript{$-$}
				& \underline{0.997}\tiny{(0.013)}\textsuperscript{$+$}
				& 0.993\tiny{(0.026)}\textsuperscript{$+$}
				& 1.002\tiny{(0.017)}
				& \textbf{1.006}\tiny{(0.021)} \\
				
				& &
				& [1.000] & [1.000] & [0.723] & [0.982] & [0.150] & [0.106] & [0.984]
				& [0.997] & [0.998] & [0.762] & [0.979] & [0.142] & [0.181] & [0.881] \\
				
				\multirow{2}{*}{$\alpha$--$C_v$}
				& \multirow{2}{*}{0.691}
				& \multirow{2}{*}{0.532}
				& 0.506\tiny{(0.021)}\textsuperscript{$-$}
				& 0.539\tiny{(0.010)}\textsuperscript{$-$}
				& 0.489\tiny{(0.030)}\textsuperscript{$-$}
				& \textbf{0.665}\tiny{(0.008)}\textsuperscript{$+$}
				& 0.647\tiny{(0.015)}\textsuperscript{$=$}
				& 0.644\tiny{(0.014)}
				& \underline{0.648}\tiny{(0.015)}
				& 0.499\tiny{(0.011)}\textsuperscript{$-$}
				& 0.527\tiny{(0.010)}\textsuperscript{$-$}
				& 0.487\tiny{(0.027)}\textsuperscript{$-$}
				& \textbf{0.656}\tiny{(0.014)}\textsuperscript{$+$}
				& 0.647\tiny{(0.017)}\textsuperscript{$=$}
				& 0.649\tiny{(0.020)}
				& \underline{0.647}\tiny{(0.015)} \\
				
				& &
				& [0.992] & [0.998] & [0.741] & [0.998] & [0.740] & [0.764] & [0.892]
				& [0.992] & [0.992] & [0.729] & [0.993] & [0.718] & [0.695] & [0.903] \\
				
				\multirow{2}{*}{$\alpha$--$\varepsilon_{\mathrm{ho}}$--$\mu$}
				& \multirow{2}{*}{0.980}
				& \multirow{2}{*}{0.713}
				& 0.373\tiny{(0.020)}\textsuperscript{$-$}
				& 0.425\tiny{(0.013)}\textsuperscript{$-$}
				& 0.392\tiny{(0.023)}\textsuperscript{$-$}
				& \textbf{0.729}\tiny{(0.018)}\textsuperscript{$+$}
				& 0.711\tiny{(0.023)}\textsuperscript{$=$}
				& 0.719\tiny{(0.024)}
				& \underline{0.712}\tiny{(0.023)}
				& 0.410\tiny{(0.022)}\textsuperscript{$-$}
				& 0.367\tiny{(0.014)}\textsuperscript{$-$}
				& 0.390\tiny{(0.047)}\textsuperscript{$-$}
				& \textbf{0.722}\tiny{(0.021)}\textsuperscript{$+$}
				& \underline{0.710}\tiny{(0.026)}\textsuperscript{$=$}
				& 0.720\tiny{(0.021)}
				& 0.705\tiny{(0.027)} \\
				
				& &
				& [0.994] & [0.999] & [0.775] & [0.925] & [0.339] & [0.341] & [0.867]
				& [0.994] & [0.993] & [0.784] & [0.956] & [0.392] & [0.355] & [0.937] \\
				
				\multirow{2}{*}{$\alpha$--$\varepsilon_{\mathrm{lu}}$--$C_v$}
				& \multirow{2}{*}{0.494}
				& \multirow{2}{*}{0.450}
				& 0.328\tiny{(0.019)}\textsuperscript{$-$}
				& 0.342\tiny{(0.010)}\textsuperscript{$-$}
				& 0.313\tiny{(0.021)}\textsuperscript{$-$}
				& \textbf{0.374}\tiny{(0.008)}\textsuperscript{$=$}
				& 0.370\tiny{(0.014)}\textsuperscript{$=$}
				& 0.370\tiny{(0.012)}
				& \underline{0.368}\tiny{(0.008)}
				& 0.318\tiny{(0.008)}\textsuperscript{$-$}
				& 0.342\tiny{(0.008)}\textsuperscript{$-$}
				& 0.306\tiny{(0.018)}\textsuperscript{$-$}
				& \underline{0.374}\tiny{(0.007)}\textsuperscript{$=$}
				& \textbf{0.376}\tiny{(0.007)}\textsuperscript{$=$}
				& 0.369\tiny{(0.013)}
				& 0.368\tiny{(0.010)} \\
				
				& &
				& [1.000] & [1.000] & [0.903] & [0.995] & [0.984] & [0.980] & [0.973]
				& [1.000] & [1.000] & [0.889] & [0.998] & [0.982] & [0.994] & [0.973] \\
				
				\multirow{2}{*}{$\Delta\varepsilon$--$\varepsilon_{\mathrm{ho}}$--$\mu$}
				& \multirow{2}{*}{0.695}
				& \multirow{2}{*}{0.669}
				& 0.463\tiny{(0.011)}\textsuperscript{$-$}
				& 0.482\tiny{(0.004)}\textsuperscript{$-$}
				& 0.459\tiny{(0.042)}\textsuperscript{$-$}
				& \textbf{0.624}\tiny{(0.019)}\textsuperscript{$+$}
				& \underline{0.616}\tiny{(0.021)}\textsuperscript{$=$}
				& 0.617\tiny{(0.021)}
				& 0.615\tiny{(0.023)}
				& 0.448\tiny{(0.018)}\textsuperscript{$-$}
				& 0.489\tiny{(0.018)}\textsuperscript{$-$}
				& 0.465\tiny{(0.027)}\textsuperscript{$-$}
				& \underline{0.627}\tiny{(0.021)}\textsuperscript{$=$}
				& 0.623\tiny{(0.026)}\textsuperscript{$=$}
				& 0.624\tiny{(0.024)}
				& \textbf{0.627}\tiny{(0.023)} \\
				
				& &
				& [1.000] & [1.000] & [0.801] & [0.978] & [0.493] & [0.511] & [0.926]
				& [1.000] & [1.000] & [0.785] & [0.970] & [0.470] & [0.419] & [0.895] \\
				
				\multirow{2}{*}{$\Delta\varepsilon$--$\varepsilon_{\mathrm{lu}}$--$C_v$}
				& \multirow{2}{*}{0.676}
				& \multirow{2}{*}{0.512}
				& 0.445\tiny{(0.024)}\textsuperscript{$-$}
				& 0.499\tiny{(0.008)}\textsuperscript{$-$}
				& 0.447\tiny{(0.021)}\textsuperscript{$-$}
				& \textbf{0.522}\tiny{(0.016)}\textsuperscript{$+$}
				& \underline{0.513}\tiny{(0.014)}\textsuperscript{$=$}
				& 0.517\tiny{(0.015)}
				& 0.510\tiny{(0.018)}
				& 0.454\tiny{(0.016)}\textsuperscript{$-$}
				& 0.506\tiny{(0.017)}\textsuperscript{$-$}
				& 0.450\tiny{(0.025)}\textsuperscript{$-$}
				& \underline{0.521}\tiny{(0.019)}\textsuperscript{$=$}
				& 0.517\tiny{(0.013)}\textsuperscript{$-$}
				& 0.522\tiny{(0.015)}
				& \textbf{0.524}\tiny{(0.016)} \\
				
				& &
				& [1.000] & [1.000] & [0.706] & [0.989] & [0.876] & [0.887] & [0.906]
				& [1.000] & [1.000] & [0.746] & [0.992] & [0.875] & [0.881] & [0.893] \\
				
				\bottomrule
			\end{tabular}
		}
	\end{table*}

	\begin{figure}[hbpt]
		\centering
		\includegraphics[width=9cm]{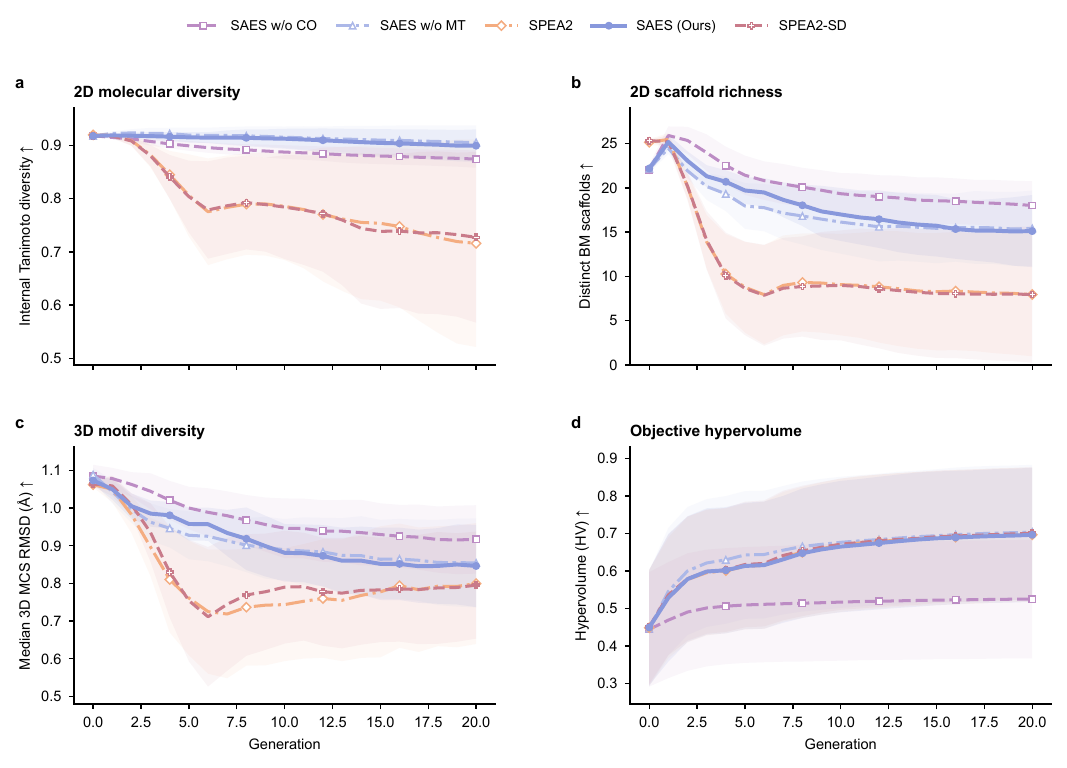}
		\caption{Structural diversity and convergence across all MOP tasks on EDM. (a) Internal Tanimoto diversity, (b) the number of distinct Bemis--Murcko scaffolds, (c) median 3D MCS RMSD, and (d) hypervolume. Lines are means across the ten tasks and shading is the standard deviation across tasks. SPEA2-SD replaces only the objective-space density term with structural distance and follows the SPEA2 diversity trajectory. SAES maintains higher structural diversity while its hypervolume stays competitive.}
		\label{fig:mop_diversity_metrics}
	\end{figure}

	Tables~\ref{tab:mop} and~\ref{tab:docking_extended} evaluate unconstrained molecular MOPs and 3D protein--ligand docking, respectively. Unlike property targeting, these tasks aim to approximate the entire Pareto front without predefined scalarization weights:
	
	\begin{equation}
		\min_{M \in \mathcal{S}} F(M) = \bigl(f_{i_1}(M), \ldots, f_{i_K}(M)\bigr)^{T}.
		\label{eq:mop_task}
	\end{equation}

	The results first demonstrate the advantage of active evolutionary search over passive sampling. Across both diffusion backbones, Top-N under the SR and SE budgets generally obtains substantially lower HV than the EGD-based evolutionary methods. As illustrated in Figure~\ref{fig:MOP_distribution}(a), unconditional samples are concentrated in the high-density region of the learned distribution and rarely reach the extreme trade-off regions. In contrast, EGD repeatedly generates new proposals from selected candidates, allowing the population to progressively expand toward the Pareto boundary.

	The results also confirm the structural diversity loss of conventional objective-space environmental selection. SPEA2 often achieves HV values comparable to SAES, but its UVR is substantially lower on most tasks. Figure~\ref{fig:MOP_distribution}(b) shows that solutions with competitive objective values can remain concentrated in a limited number of structural regions. Objective-space crowding therefore does not guarantee diversity in molecular structure space.

	SAES alleviates this mismatch by incorporating both topological and geometric distances into environmental selection. It maintains HV performance comparable to SPEA2 while retaining substantially higher structural diversity. As shown in Figure~\ref{fig:MOP_distribution}(e), SAES covers the Pareto front with structurally distinct solutions rather than repeatedly selecting minor variants of similar molecules. These results show that structural diversity can be preserved without sacrificing objective-space convergence.
	
	Figure~\ref{fig:mop_diversity_metrics} and the SPEA2-SD column of Table~\ref{tab:mop} show why this behavior cannot be obtained by simply replacing objective-space diversity with structural diversity. In SPEA2-SD the dominance term still always outweighs the structural term, and objective-space diversity is dropped rather than compared with structural diversity. Its 2D diversity, scaffold count, and 3D motif diversity therefore fall with SPEA2, and its UVR stays at the SPEA2 level, even though HV remains comparable. Rather than unconditionally prioritizing convergence, SAES requires that among solutions with comparable convergence, structural diversity must take precedence over objective-space diversity. This is a tradeoff between convergence and structural diversity, rather than a rule in which convergence always dominates structural diversity.
	
	The ablation results further clarify the roles of the EGD operators. Removing crossover substantially weakens HV convergence, while removing mutation has a smaller effect on most MOP tasks. This indicates that crossover mainly supports global exploration, whereas mutation provides local refinement. Their combination with SAES enables the population to expand the Pareto front while maintaining structural diversity.

\begin{figure}[hbpt]
		\centering
		\includegraphics[width=9cm]{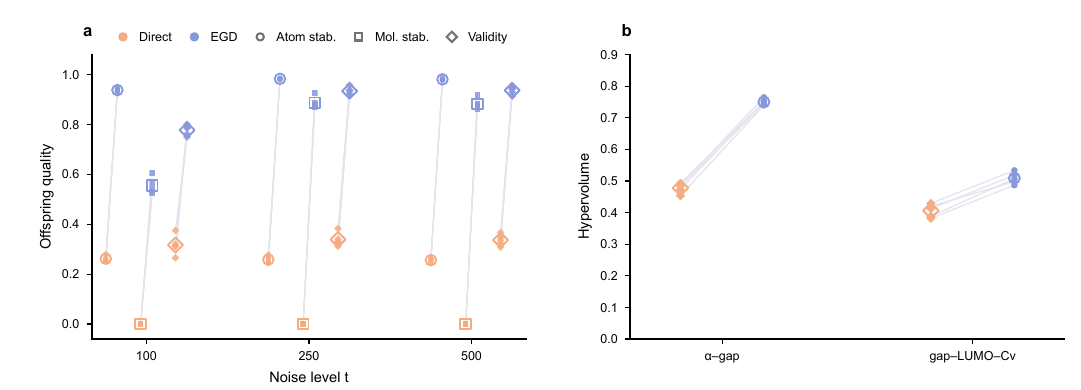}
		\caption{Comparison between direct crossover in the GeoLDM latent space and EGD in the noise space. (a) Chemical validity and stability of unscreened offspring across three noise levels. (b) Hypervolume after search on the $\alpha$--gap and gap--LUMO--$C_v$ tasks. Paired seeds are connected by lines, and open markers indicate seed means.}
		\label{fig:latent_recombination}
	\end{figure}

	To study crossover in the noise space, we compare EGD with direct crossover in the latent variables of GeoLDM (i.e., the VAE part of the same model, whose decoder maps standard Gaussian noise to a molecular solution). Both operators exchange the same atomic subsets of the same parents; however, direct crossover decodes the recombined latent state, whereas EGD noises that state and completes it via reverse diffusion. Figure~\ref{fig:latent_recombination}(a) shows that direct latent crossover produces chemically unreasonable molecules, whereas EGD raises reasonableness substantially. This occurs because a 3D molecule is a complex, variable-sized, and equivariant object, and the VAE does not form a continuous latent space in which direct crossover remains reasonable; consequently, the recombined codes readily decode to invalid structures. EGD avoids this failure by letting reverse diffusion complete the recombined state. As a result, these reasonable candidates make the subsequent search effective, allowing EGD to achieve better convergence than direct latent crossover, as shown in Figure~\ref{fig:latent_recombination}(b).

	Docking uses a different objective vector on each protein pocket. The minimized scores are the Vina energy, the negated QED, and the synthetic-accessibility score:
	\begin{equation}
		\min_{M} F(M) = \bigl(E_{\mathrm{Vina}}(M),\, -\mathrm{QED}(M),\, \mathrm{SA}(M)\bigr).
		\label{eq:dock_task}
	\end{equation}
	A more negative Vina score, a larger QED, and a smaller SA score are preferred, so QED enters with a minus.

		\begin{table}[htbp]
		\centering
		\caption{Extended HV ($\uparrow$) and UVR ($\uparrow$) results on the 3D Protein-Ligand Docking MOP task. Values are formatted as \textbf{HV(Std) - UVR}. }
		\label{tab:docking_extended}
		\resizebox{\columnwidth}{!}{
			\begin{tabular}{ll cc}
				\toprule
				\multirow{2}{*}{\textbf{Target}} & \textbf{Top-N} (SE) & \multicolumn{2}{c}{\textbf{Evolutionary Framework}} \\
				\cmidrule(lr){2-2} \cmidrule(lr){3-4}
				& Unconditional & SPEA2 & \textbf{SAES (Ours)} \\
				\midrule
				\multicolumn{4}{c}{\textit{Backbone: GeoLDM}} \\
				\midrule
				4tqr & 0.718{\tiny (.080)}\textsuperscript{$-$}- 1.00 & \textbf{0.798}{\tiny (.095)}\textsuperscript{$=$}- 0.84 & \underline{0.793}{\tiny (.081)}- 0.97 \\
				1k9t & \textbf{0.339}{\tiny (.038)}\textsuperscript{$=$}- 0.60 & \underline{0.338}{\tiny (.028)}\textsuperscript{$=$}- 0.47 & 0.328{\tiny (.018)}- 0.64 \\
				1ai4 & 0.458{\tiny (.033)}\textsuperscript{$-$}- 0.95 & \underline{0.481}{\tiny (.043)}\textsuperscript{$-$}- 0.71 & \textbf{0.490}{\tiny (.049)}- 0.92 \\
				5mgl & 0.485{\tiny (.039)}\textsuperscript{$-$}- 0.91 & \underline{0.517}{\tiny (.043)}\textsuperscript{$-$}- 0.65 & \textbf{0.519}{\tiny (.038)}- 0.87 \\
				3af2 & 0.565{\tiny (.073)}\textsuperscript{$-$}- 1.00 & \textbf{0.748}{\tiny (.125)}\textsuperscript{$=$}- 0.92 & \underline{0.741}{\tiny (.075)}- 0.94 \\
				2rma & 0.554{\tiny (.051)}\textsuperscript{$-$}- 1.00 & \textbf{0.585}{\tiny (.075)}\textsuperscript{$=$}- 0.88 & \underline{0.581}{\tiny (.047)}- 0.97 \\
				2pc8 & 0.828{\tiny (.065)}\textsuperscript{$-$}- 0.97 & \underline{0.800}{\tiny (.066)}\textsuperscript{$-$}- 0.75 & \textbf{0.835}{\tiny (.082)}- 0.92 \\
				1r1h & 0.596{\tiny (.059)}\textsuperscript{$-$}- 1.00 & \underline{0.757}{\tiny (.096)}\textsuperscript{$-$}- 0.78 & \textbf{0.771}{\tiny (.061)}- 0.95 \\
				1jn2 & 0.684{\tiny (.067)}\textsuperscript{$-$}- 1.00 & \textbf{0.803}{\tiny (.119)}\textsuperscript{$=$}- 0.84 & \underline{0.799}{\tiny (.091)}- 0.96 \\
				4zfa & 0.668{\tiny (.060)}\textsuperscript{$-$}- 0.99 & \underline{0.772}{\tiny (.110)}\textsuperscript{$-$}- 0.81 & \textbf{0.789}{\tiny (.051)}- 0.93 \\
				\midrule
				\multicolumn{4}{c}{\textit{Backbone: EDM}} \\
				\midrule
				4tqr & 0.700{\tiny (.067)}\textsuperscript{$-$}- 1.00 & \underline{0.815}{\tiny (.067)}\textsuperscript{$-$}- 0.88 & \textbf{0.839}{\tiny (.100)}- 0.96 \\
				1k9t & 0.440{\tiny (.060)}\textsuperscript{$-$}- 0.89 & \underline{0.463}{\tiny (.065)}\textsuperscript{$-$}- 0.48 & \textbf{0.473}{\tiny (.042)}- 0.92 \\
				1ai4 & 0.513{\tiny (.031)}\textsuperscript{$-$}- 0.97 & \underline{0.544}{\tiny (.041)}\textsuperscript{$=$}- 0.73 & \textbf{0.545}{\tiny (.038)}- 0.95 \\
				5mgl & 0.577{\tiny (.037)}\textsuperscript{$-$}- 0.94 & \underline{0.599}{\tiny (.057)}\textsuperscript{$-$}- 0.72 & \textbf{0.616}{\tiny (.044)}- 0.92 \\
				3af2 & 0.504{\tiny (.064)}\textsuperscript{$-$}- 1.00 & \underline{0.710}{\tiny (.106)}\textsuperscript{$=$}- 0.86 & \textbf{0.716}{\tiny (.103)}- 0.98 \\
				2rma & 0.590{\tiny (.060)}\textsuperscript{$-$}- 1.00 & \textbf{0.614}{\tiny (.077)}\textsuperscript{$=$}- 0.89 & \underline{0.613}{\tiny (.067)}- 0.98 \\
				2pc8 & 0.845{\tiny (.045)}\textsuperscript{$-$}- 0.98 & \underline{0.922}{\tiny (.090)}\textsuperscript{$=$}- 0.84 & \textbf{0.923}{\tiny (.078)}- 0.92 \\
				1r1h & 0.568{\tiny (.078)}\textsuperscript{$-$}- 1.00 & \underline{0.704}{\tiny (.091)}\textsuperscript{$-$}- 0.89 & \textbf{0.709}{\tiny (.076)}- 0.98 \\
				1jn2 & 0.679{\tiny (.063)}\textsuperscript{$-$}- 1.00 & \textbf{0.812}{\tiny (.103)}\textsuperscript{$=$}- 0.89 & \underline{0.810}{\tiny (.104)}- 0.97 \\
				4zfa & 0.711{\tiny (.080)}\textsuperscript{$-$}- 1.00 & \underline{0.779}{\tiny (.112)}\textsuperscript{$-$}- 0.87 & \textbf{0.787}{\tiny (.090)}- 0.97 \\
				\bottomrule
			\end{tabular}
		}
	\end{table}
	
	To examine whether these advantages extend to a more practical molecular design setting, we further evaluate the objective in \eqref{eq:dock_task} on 3D protein--ligand docking across 10 protein pockets. Unlike most existing SBDD generative models, our diffusion model is trained only on the general-purpose GEOM-Drugs dataset and has never seen protein-pocket information. This setting is practically important because many newly identified or less-studied protein targets do not have enough known ligands or protein--ligand complexes to train a target-specific generative model. Reusing a general molecular generator would therefore allow molecular design to start directly from a pretrained chemical prior and adapt to a new pocket through optimization, without collecting a new SBDD training set or retraining the generative model.
	
	For this reason, we do not directly compare with SBDD models trained on protein--ligand complexes, since such a comparison would confound task-specific training with optimization capability. Instead, Top-N, SPEA2, and SAES use the same GEOM-Drugs pretrained diffusion backbone and differ only in their search strategies. This controlled setting evaluates whether evolutionary optimization can transfer a non-SBDD molecular prior to previously unseen protein pockets.
	
	As reported in Table~\ref{tab:docking_extended}, Top-N generally maintains a high UVR but produces lower HV because passive sampling cannot effectively approach the Pareto boundary. SPEA2 substantially improves HV but reduces structural diversity, whereas SAES achieves statistically comparable or better HV on most targets while consistently maintaining a higher UVR than SPEA2. These results demonstrate that EGD and SAES can directly adapt a general-purpose 3D molecular diffusion model to structure-based drug design while maintaining both optimization performance and molecular diversity.

	\subsection{Results on CMOPs}

	Each CMOP task minimizes the property vector named in its row and requires both specified fragments to be retained:
	\begin{equation}
		\min_{M \in \mathcal{S}} F(M)
		\quad \text{s.t.} \quad
		CV_{\phi}(M) = 0,
		\label{eq:cmop_task}
	\end{equation}
	where $CV_{\phi}$ is the fragment-violation score in \eqref{eq:fragment_cv}. 

	\begin{table*}[htbp]
		\centering
		\caption{Comprehensive HV ($\uparrow$) and [FR $\uparrow$ / UFR $\uparrow$] results on the CMOP multi-fragment generation task.}
		\label{tab:cmop_all}
		\resizebox{\textwidth}{!}{
			\begin{tabular}{ll cccc ccc |ll cccc ccc}
				\toprule
				\multirow{2}{*}{\textbf{Task}} & \textbf{Top-N} & \multicolumn{3}{c}{\textbf{Traditional CMOEAs}} & \multicolumn{3}{c}{\textbf{Ablation \& DEMO}} & \multirow{2}{*}{\textbf{Task}} & \textbf{Top-N} & \multicolumn{3}{c}{\textbf{Traditional CMOEAs}} & \multicolumn{3}{c}{\textbf{Ablation \& DEMO}} \\
				\cmidrule(lr){2-2} \cmidrule(lr){3-5} \cmidrule(lr){6-8} \cmidrule(lr){10-10} \cmidrule(lr){11-13} \cmidrule(lr){14-16}
				& (SE) & SPEA2 & CCMO & CMOEA-CD & w/o Pop B & Only Pop C & \textbf{Ours} & & (SE) & SPEA2 & CCMO & CMOEA-CD & w/o Pop B & Only Pop C & \textbf{Ours} \\
				\midrule
				\multicolumn{8}{c}{\textit{Backbone: GeoLDM}} & \multicolumn{8}{c}{\textit{Backbone: EDM}} \\
				\midrule
				\multirow{2}{*}{$\alpha$---$\Delta\varepsilon$} 
				& 0.035\tiny{(.085)}\textsuperscript{$-$} & 0.352\tiny{(.023)}\textsuperscript{$-$} & 0.343\tiny{(.028)}\textsuperscript{$-$} & 0.338\tiny{(.033)}\textsuperscript{$-$} & \textbf{0.399}\tiny{(.042)}\textsuperscript{$=$} & 0.386\tiny{(.062)}\textsuperscript{$=$} & \underline{0.394}\tiny{(.036)} &
				\multirow{2}{*}{$\alpha$---$\Delta\varepsilon$} 
				& 0.104\tiny{(.014)}\textsuperscript{$-$} & \underline{0.141}\tiny{(.137)}\textsuperscript{$=$} & 0.111\tiny{(.133)}\textsuperscript{$-$} & 0.093\tiny{(.012)}\textsuperscript{$-$} & 0.091\tiny{(.143)}\textsuperscript{$-$} & 0.182\tiny{(.136)}\textsuperscript{$=$} & \textbf{0.143}\tiny{(.147)} \\
				& [0.04/0.01] & [1.00/0.13] & [1.00/0.17] & [1.00/0.18] & [1.00/0.46] & [0.15/0.11] & [1.00/0.53] &
				& [0.01/0.01] & [0.47/0.05] & [0.31/0.03] & [0.31/0.02] & [0.24/0.03] & [0.03/0.03] & [0.44/0.03] \\
				
				\multirow{2}{*}{-$\Delta\varepsilon$--$\varepsilon_{\text{ho}}$} 
				& 0\tiny{(.0)}\textsuperscript{$-$} & 0.015\tiny{(.069)}\textsuperscript{$-$} & 0.011\tiny{(.048)}\textsuperscript{$-$} & \textbf{0.025}\tiny{(.077)}\textsuperscript{$=$} & \underline{0.024}\tiny{(.081)}\textsuperscript{$=$} & 0\tiny{(.0)}\textsuperscript{$-$} & \textbf{0.025}\tiny{(.067)} &
				\multirow{2}{*}{-$\Delta\varepsilon$--$\varepsilon_{\text{ho}}$} 
				& 0.251\tiny{(.068)}\textsuperscript{$-$} & 0.297\tiny{(.084)}\textsuperscript{$-$} & 0.316\tiny{(.038)}\textsuperscript{$-$} & 0.297\tiny{(.074)}\textsuperscript{$-$} & \textbf{0.358}\tiny{(.047)}\textsuperscript{$=$} & \textbf{0.358}\tiny{(.044)}\textsuperscript{$=$} & \underline{0.356}\tiny{(.044)} \\
				& [0.00/0.00] & [0.05/0.01] & [0.05/0.01] & [0.10/0.01] & [0.01/0.01] & [0.00/0.00] & [0.10/0.01] &
				& [0.12/0.11] & [0.95/0.11] & [0.96/0.12] & [0.95/0.15] & [1.00/0.34] & [0.11/0.11] & [1.00/0.42] \\
				
				\multirow{2}{*}{$\varepsilon_{\text{ho}}$--$\varepsilon_{\text{lu}}$} 
				& 0.019\tiny{(.059)}\textsuperscript{$-$} & 0.161\tiny{(.057)}\textsuperscript{$-$} & 0.180\tiny{(.071)}\textsuperscript{$-$} & 0.150\tiny{(.074)}\textsuperscript{$-$} & \underline{0.246}\tiny{(.030)}\textsuperscript{$=$} & 0.259\tiny{(.043)}\textsuperscript{$=$} & \textbf{0.263}\tiny{(.044)} &
				\multirow{2}{*}{$\varepsilon_{\text{ho}}$--$\varepsilon_{\text{lu}}$} 
				& 0.205\tiny{(.050)}\textsuperscript{$-$} & 0.294\tiny{(.038)}\textsuperscript{$-$} & 0.291\tiny{(.033)}\textsuperscript{$-$} & 0.278\tiny{(.057)}\textsuperscript{$-$} & \underline{0.480}\tiny{(.033)}\textsuperscript{$=$} & 0.500\tiny{(.046)}\textsuperscript{$=$} & \textbf{0.502}\tiny{(.042)} \\
				& [0.01/0.01] & [0.93/0.08] & [0.90/0.07] & [0.81/0.11] & [1.00/0.48] & [0.16/0.11] & [1.00/0.54] &
				& [0.15/0.12] & [1.00/0.19] & [1.00/0.19] & [1.00/0.23] & [1.00/0.74] & [0.41/0.32] & [1.00/0.79] \\
				
				\multirow{2}{*}{$\varepsilon_{\text{lu}}$--$\mu$} 
				& 0\tiny{(.0)}\textsuperscript{$-$} & 0.374\tiny{(.176)}\textsuperscript{$-$} & 0.355\tiny{(.159)}\textsuperscript{$-$} & 0.341\tiny{(.158)}\textsuperscript{$-$} & \underline{0.617}\tiny{(.069)}\textsuperscript{$=$} & 0.604\tiny{(.072)}\textsuperscript{$=$} & \textbf{0.626}\tiny{(.063)} &
				\multirow{2}{*}{$\varepsilon_{\text{lu}}$--$\mu$} 
				& 0\tiny{(.0)}\textsuperscript{$-$} & 0.041\tiny{(.125)}\textsuperscript{$-$} & 0.092\tiny{(.163)}\textsuperscript{$-$} & 0.038\tiny{(.110)}\textsuperscript{$-$} & \underline{0.589}\tiny{(.147)}\textsuperscript{$=$} & 0.547\tiny{(.125)}\textsuperscript{$=$} & \textbf{0.606}\tiny{(.112)} \\
				& [0.00/0.00] & [0.80/0.08] & [0.82/0.08] & [0.82/0.10] & [1.00/0.45] & [0.15/0.13] & [1.00/0.48] &
				& [0.00/0.00] & [0.10/0.01] & [0.21/0.02] & [0.10/0.01] & [0.95/0.26] & [0.09/0.08] & [0.97/0.28] \\
				
				\multirow{2}{*}{$\mu$--$C_v$} 
				& 0\tiny{(.0)}\textsuperscript{$-$} & 0.732\tiny{(.059)}\textsuperscript{$-$} & 0.762\tiny{(.188)}\textsuperscript{$=$} & 0.748\tiny{(.071)}\textsuperscript{$=$} & \underline{0.760}\tiny{(.056)}\textsuperscript{$=$} & 0.721\tiny{(.035)}\textsuperscript{$-$} & \textbf{0.771}\tiny{(.047)} &
				\multirow{2}{*}{$\mu$--$C_v$} 
				& 0\tiny{(.0)}\textsuperscript{$-$} & 0\tiny{(.0)}\textsuperscript{$-$} & 0\tiny{(.0)}\textsuperscript{$-$} & 0\tiny{(.0)}\textsuperscript{$-$} & 0\tiny{(.0)}\textsuperscript{$-$} & \textbf{0.080}\tiny{(.191)}\textsuperscript{$=$} & \underline{0.060}\tiny{(.146)} \\
				& [0.00/0.00] & [1.00/0.12] & [0.92/0.11] & [0.99/0.14] & [1.00/0.38] & [0.19/0.12] & [1.00/0.44] &
				& [0.00/0.00] & [0.00/0.00] & [0.00/0.00] & [0.00/0.00] & [0.00/0.00] & [0.01/0.01] & [0.15/0.01] \\
				
				\multirow{2}{*}{$C_v$--$\alpha$} 
				& 0.103\tiny{(.141)}\textsuperscript{$=$} & 0\tiny{(.0)}\textsuperscript{$-$} & 0\tiny{(.0)}\textsuperscript{$-$} & 0\tiny{(.0)}\textsuperscript{$-$} & 0\tiny{(.0)}\textsuperscript{$-$} & 0.011\tiny{(.051)}\textsuperscript{$-$} & \textbf{0.123}\tiny{(.141)} &
				\multirow{2}{*}{$C_v$--$\alpha$} 
				& 0.358\tiny{(.016)}\textsuperscript{$-$} & \underline{0.423}\tiny{(.018)}\textsuperscript{$-$} & 0.130\tiny{(.017)}\textsuperscript{$-$} & 0.420\tiny{(.019)}\textsuperscript{$-$} & 0.461\tiny{(.012)}\textsuperscript{$=$} & \underline{0.478}\tiny{(.023)}\textsuperscript{$=$} & \textbf{0.491}\tiny{(.025)} \\
				& [0.01/0.01] & [0.00/0.00] & [0.00/0.00] & [0.00/0.00] & [0.00/0.00] & [0.00/0.00] & [0.11/0.02] &
				& [0.26/0.15] & [1.00/0.31] & [1.00/0.33] & [1.00/0.35] & [0.99/0.45] & [0.28/0.26] & [1.00/0.73] \\
				
				\multirow{2}{*}{$\alpha$--$\varepsilon_{\text{ho}}$--$\mu$} 
				& 0\tiny{(.0)}\textsuperscript{$-$} & 0.122\tiny{(.055)}\textsuperscript{$-$} & 0.087\tiny{(.080)}\textsuperscript{$-$} & 0.095\tiny{(.071)}\textsuperscript{$-$} & \underline{0.218}\tiny{(.023)}\textsuperscript{$=$} & 0.148\tiny{(.093)}\textsuperscript{$-$} & \textbf{0.237}\tiny{(.020)} &
				\multirow{2}{*}{$\alpha$--$\varepsilon_{\text{ho}}$--$\mu$} 
				& 0.141\tiny{(.061)}\textsuperscript{$-$} & 0.089\tiny{(.090)}\textsuperscript{$-$} & 0.095\tiny{(.096)}\textsuperscript{$-$} & 0.061\tiny{(.085)}\textsuperscript{$-$} & \underline{0.226}\tiny{(.029)}\textsuperscript{$=$} & 0.133\tiny{(.079)}\textsuperscript{$-$} & \textbf{0.244}\tiny{(.034)} \\
				& [0.00/0.00] & [0.09/0.08] & [0.55/0.05] & [0.57/0.05] & [1.00/0.39] & [0.04/0.04] & [1.00/0.44] &
				& [0.03/0.03] & [0.45/0.05] & [0.50/0.06] & [0.35/0.03] & [1.00/0.44] & [0.04/0.04] & [1.00/0.34] \\
				
				\multirow{2}{*}{$\alpha$--$\varepsilon_{\text{lu}}$--$C_v$} 
				& 0\tiny{(.0)}\textsuperscript{$-$} & 0.019\tiny{(.047)}\textsuperscript{$-$} & 0.012\tiny{(.037)}\textsuperscript{$-$} & 0.022\tiny{(.053)}\textsuperscript{$-$} & \textbf{0.085}\tiny{(.081)}\textsuperscript{$=$} & 0.008\tiny{(.026)}\textsuperscript{$-$} & \underline{0.045}\tiny{(.062)} &
				\multirow{2}{*}{$\alpha$--$\varepsilon_{\text{lu}}$--$C_v$} 
				& 0.024\tiny{(.042)}\textsuperscript{$-$} & 0.102\tiny{(.081)}\textsuperscript{$=$} & 0.060\tiny{(.078)}\textsuperscript{$-$} & 0.072\tiny{(.084)}\textsuperscript{$-$} & \underline{0.103}\tiny{(.073)}\textsuperscript{$=$} & 0\tiny{(.0)}\textsuperscript{$-$} & \textbf{0.111}\tiny{(.081)} \\
				& [0.00/0.00] & [0.11/0.02] & [0.21/0.10] & [0.11/0.01] & [0.49/0.06] & [0.01/0.01] & [0.14/0.02] &
				& [0.01/0.01] & [0.56/0.11] & [0.38/0.04] & [0.41/0.07] & [0.66/0.07] & [0.00/0.00] & [0.70/0.09] \\
				
				\multirow{2}{*}{-$\Delta\varepsilon$--$\varepsilon_{\text{ho}}$--$\mu$} 
				& 0\tiny{(.0)}\textsuperscript{$-$} & 0.160\tiny{(.132)}\textsuperscript{$-$} & 0.181\tiny{(.133)}\textsuperscript{$-$} & 0.161\tiny{(.133)}\textsuperscript{$-$} & \underline{0.202}\tiny{(.081)}\textsuperscript{$=$} & 0.125\tiny{(.012)}\textsuperscript{$-$} & \textbf{0.220}\tiny{(.111)} &
				\multirow{2}{*}{-$\Delta\varepsilon$--$\varepsilon_{\text{ho}}$--$\mu$} 
				& 0.197\tiny{(.055)}\textsuperscript{$-$} & 0.278\tiny{(.035)}\textsuperscript{$-$} & 0.279\tiny{(.028)}\textsuperscript{$-$} & 0.254\tiny{(.026)}\textsuperscript{$-$} & \underline{0.308}\tiny{(.034)}\textsuperscript{$=$} & 0.096\tiny{(.108)}\textsuperscript{$-$} & \textbf{0.323}\tiny{(.037)} \\
				& [0.00/0.00] & [0.58/0.18] & [0.65/0.19] & [0.56/0.21] & [0.20/0.20] & [0.03/0.03] & [0.80/0.26] &
				& [0.08/0.08] & [1.00/0.21] & [1.00/0.23] & [0.96/0.20] & [1.00/0.55] & [0.02/0.02] & [1.00/0.52] \\
				
				\multirow{2}{*}{-$\Delta\varepsilon$--$\varepsilon_{\text{lu}}$--$C_v$} 
				& 0.011\tiny{(.051)}\textsuperscript{$-$} & 0.282\tiny{(.077)}\textsuperscript{$=$} & \underline{0.289}\tiny{(.060)}\textsuperscript{$=$} & 0.287\tiny{(.067)}\textsuperscript{$=$} & 0.291\tiny{(.060)}\textsuperscript{$=$} & 0.260\tiny{(.077)}\textsuperscript{$-$} & \textbf{0.306}\tiny{(.050)} &
				\multirow{2}{*}{-$\Delta\varepsilon$--$\varepsilon_{\text{lu}}$--$C_v$} 
				& 0.307\tiny{(.035)}\textsuperscript{$-$} & 0.385\tiny{(.026)}\textsuperscript{$=$} & \underline{0.388}\tiny{(.025)}\textsuperscript{$=$} & \textbf{0.392}\tiny{(.020)}\textsuperscript{$=$} & 0.389\tiny{(.029)}\textsuperscript{$=$} & 0.137\tiny{(.094)}\textsuperscript{$-$} & 0.387\tiny{(.041)} \\
				& [0.01/0.01] & [0.95/0.26] & [0.96/0.22] & [0.95/0.24] & [1.00/0.38] & [0.11/0.10] & [1.00/0.37] &
				& [0.50/0.45] & [1.00/0.40] & [1.00/0.46] & [1.00/0.47] & [1.00/0.56] & [0.04/0.04] & [1.00/0.64] \\
				\bottomrule
			\end{tabular}
		}
	\end{table*}

The multi-fragment CMOP is designed to mimic a common setting in fragment-based molecular design, where several functional fragments are known in advance and must be simultaneously retained within a complete molecule with desirable physicochemical properties. Using two fragments provides a minimal but non-trivial setting: their incorporation pattern, connectivity to the remaining molecular structure, and resulting 3D conformation are not predefined, while both fragment motifs must be retained during optimization.

Table~\ref{tab:cmop_all} reports the results of 10 fragment-constrained property combinations using both GeoLDM and EDM, resulting in 20 independent CMOP tasks. To isolate the influence of the optimization framework, SPEA2-CDP, CCMO, and CMOEA-CD all use EGD as their molecular variation operator.

Although EGD enables the traditional CMOEAs to generate feasible molecules, their UFR is substantially lower than their FR on most tasks. This gap indicates that many feasible candidates are structurally repeated or differ only by minor modifications. Their objective- and constraint-space selection strategies therefore tend to repeatedly exploit a limited set of feasible structures rather than preserve alternative structural pathways.

Table~\ref{tab:cmop_all} shows that DEMO generally achieves the best or statistically competitive HV while maintaining a markedly higher UFR than the traditional CMOEAs. These results indicate that preserving structurally distinct feasible molecules not only improves the diversity of the final solution set but also supports continued Pareto-front expansion under stringent fragment constraints.

\begin{figure}[hbpt]
	\centering
	\includegraphics[width=\linewidth]{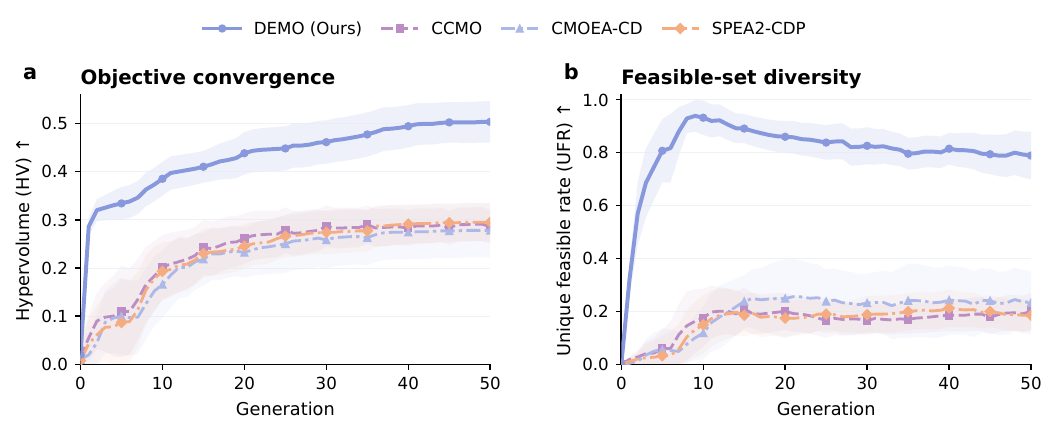}
	\caption{UFR and HV on the $\varepsilon_{\mathrm{ho}}$--$\varepsilon_{\mathrm{lu}}$ task.}
	\label{fig:cmop_curves}
\end{figure}

Figure~\ref{fig:cmop_curves} further illustrates the relationship between structural diversity and convergence on the $\varepsilon_{\mathrm{ho}}$--$\varepsilon_{\mathrm{lu}}$ task. The UFR of the traditional CMOEAs decreases rapidly as feasible variants of a limited number of structures accumulate in the population. Their HV improvement subsequently slows because structurally redundant populations provide few alternative directions for further optimization. In contrast, DEMO maintains a higher UFR, allowing structurally distinct feasible molecules to remain available for continued property refinement.

\begin{figure*}[htbp]
	\centering
	\vspace{-3em}
	\includegraphics[width=\textwidth]{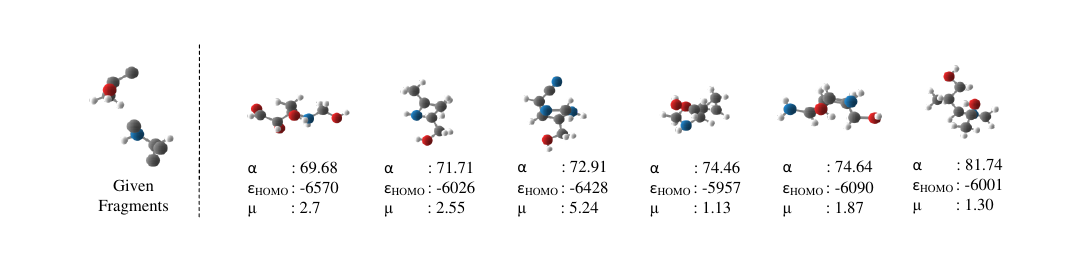} 
	
	\vspace{-3em}

	\includegraphics[width=\textwidth]{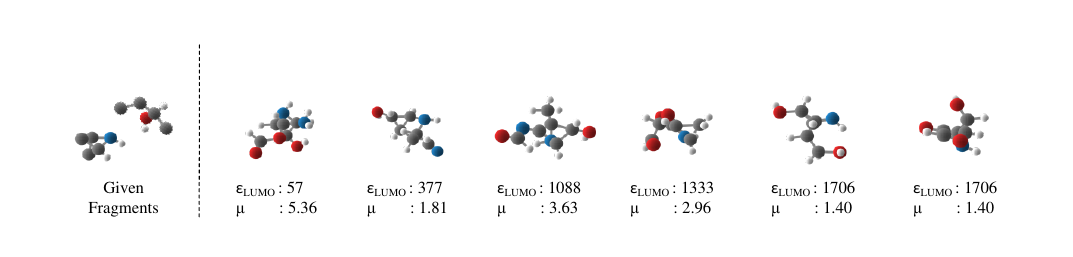}

	\vspace{-3em}
	
	\includegraphics[width=\textwidth]{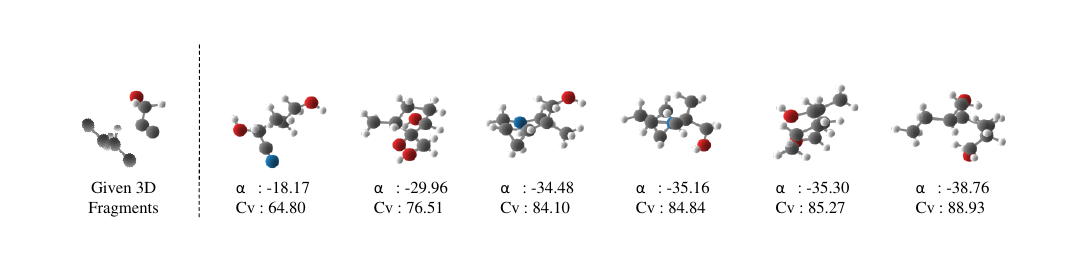}

	\caption{Additional CMOP visualization results across different property combinations and diffusion backbones.   DEMO optimizes the conflicting properties while retaining the specified target fragment substructures. (1) $\alpha$, $\varepsilon_{\mathrm{ho}}$, and $\mu$ with GeoLDM; (2) $\varepsilon_{\mathrm{lu}}$ and $\mu$ with EDM; }
	\label{fig:cmop_visualizations_si}
\end{figure*}

The ablation study further clarifies the roles of the three populations. The Only $\mathcal{P}_C$ variant generally obtains lower FR and HV because it directly searches for feasible and high-quality molecules without preserving infeasible intermediates. It therefore struggles to cross the sparse structural regions separating partially assembled and fully feasible molecules. Removing $\mathcal{P}_B$ also reduces UFR on most tasks. Without a population dedicated to progressive fragment incorporation, candidates generated by global exploration must directly enter feasible-solution refinement, causing the search to repeatedly exploit a limited set of successfully assembled structures. The complete tri-population architecture coordinates global structural exploration in $\mathcal{P}_A$, progressive fragment incorporation in $\mathcal{P}_B$, and conformation refinement in $\mathcal{P}_C$.

\begin{figure}[hbpt]
	\centering
	\includegraphics[width=\columnwidth]{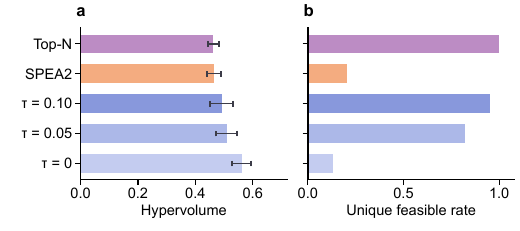}
	\caption{Ablation of the SAES structural-distance threshold $\tau$ on the EDM HOMO--LUMO fragment-constrained task. (a) Final hypervolume and (b) unique feasible rate. Error bars denote the sample standard deviation. }
	\label{fig:tau_ablation}
\end{figure}

Figure~\ref{fig:tau_ablation} ablates the structural-distance threshold \(\tau\) on the EDM \(\varepsilon_{\text{ho}}\)--\(\varepsilon_{\text{lu}}\) fragment-constrained task.  While all methods achieve a final fragment feasibility rate of 1, they exhibit a clear trade-off between objective convergence and structural diversity. Removing the threshold (\(\tau=0\)) yields the highest hypervolume but causes the unique feasible rate to collapse, as the population fills with structurally redundant variants. Increasing \(\tau\) enforces structural distinctness, progressively raising the unique feasible rate at a slight cost to hypervolume. The setting \(\tau=0.05\) strikes a balance between these two aspects and is thus adopted for the main experiments. Notably, even at \(\tau=0\), SAES differs from SPEA2: SAES still computes its density estimation in the structural space rather than the objective space, yet without the hard threshold, it cannot prevent structural redundancy. In contrast, SPEA2, which performs selection purely in the objective space, and unconditional Top-\(N\), which does not apply evolutionary variation, both fall short of this balance.

\begin{figure}[hbpt]
	\centering
	\includegraphics[width=\columnwidth]{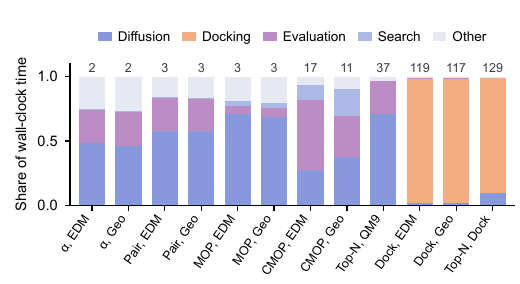}
	\caption{Share of the wall-clock time of one run, averaged over five repeats; numbers above the bars are rounded mean minutes per run. Task abbreviations: Pair, MOP, and CMOP are $\varepsilon_{\mathrm{ho}}$--$\varepsilon_{\mathrm{lu}}$ tasks; Dock is docking; Geo is GeoLDM. Top-N, QM9 denotes unconditional Top-$N$ sampling on the $\varepsilon_{\mathrm{ho}}$--$\varepsilon_{\mathrm{lu}}$ fragment-constrained task with the EDM backbone, and Top-N, Dock its counterpart on docking. Each bar is divided into Diffusion (reverse diffusion), Docking (pose scoring), Evaluation (property, chemical, and fragment checks), Search (crossover, noising, selection, and adaptive scheduling), and Other (model loading, initial sampling, and unlisted overhead).}
	\label{fig:runtime}
\end{figure}

Figure~\ref{fig:runtime} shows where the wall-clock time goes. EDM and GeoLDM follow the same pattern. On property targeting and unconstrained optimization, reverse diffusion is the main cost, and each generation stays short. Fragment constraints move time toward subgraph matching, so evaluation grows and the run becomes longer. Docking changes the balance: pose scoring dominates, and reverse diffusion shrinks to a thin slice of a much longer run. Unconditional Top-$N$ on the $\varepsilon_{\mathrm{ho}}$--$\varepsilon_{\mathrm{lu}}$ fragment-constrained task (EDM backbone) takes longer than that evolutionary run. Its time goes into drawing and screening a large sample, not into search, and every Top-$N$ sample is denoised over the full trajectory from $T$ to $0$, whereas EGD starts reverse diffusion from the intermediate timestep $t'$, so each EGD offspring is cheaper to generate. The adaptive scheduler does not drive any of these costs. It adds no extra denoising calls, and the search band that contains it stays narrow because the scheduler itself takes only a few seconds.

	\section{Conclusion and Future Work}
	
	In summary, DEMO combines the search capability of evolutionary algorithms with the generative capability of pretrained 3D diffusion models for constrained multi-objective molecular design. EGD enables evolutionary variation on variable-sized 3D molecules without retraining the diffusion model, while SAES mitigates structural diversity loss caused by objective-space selection. The tri-population architecture further coordinates global structural exploration, progressive fragment assembly, and physicochemical refinement. Experimental results across property targeting, unconstrained MOPs, protein--ligand docking, and fragment-constrained CMOPs demonstrate that DEMO can improve objective convergence while maintaining structurally diverse solutions. These findings establish DEMO as an effective zero-shot framework for constrained multi-objective 3D molecular discovery.
	
	Several directions remain for future research. First, the exploration range of EGD is still influenced by the distribution learned by the pretrained diffusion model. A closed-loop framework could incrementally fine-tune the model using newly discovered high-quality molecules, thereby expanding the accessible chemical space beyond the original training distribution. Second, human- or LLM-in-the-loop guidance could improve the controllability of the evolutionary process. Domain knowledge could be used to select candidate fragments, determine molecular regions for variation, or adapt the exploration strategy, enabling more targeted and interpretable molecular optimization. Third, the property-predictor and docking evaluations used in this work are computationally mild compared with quantum-chemical, molecular-dynamics, free-energy, or experimental measurements. Surrogate-assisted environmental selection and Bayesian active-learning acquisition could prioritize which candidates receive such expensive evaluations under a limited budget, extending DEMO to costlier oracles.

	\bibliography{reference}

@inproceedings{mudm,
  title={Training-free multi-objective diffusion model for 3d molecule generation},
  author={Han, Xu and Shan, Caihua and Shen, Yifei and Xu, Can and Yang, Han and Li, Xiang and Li, Dongsheng},
  booktitle={The Twelfth International Conference on Learning Representations},
  year={2023}
}

@inproceedings{edm,
  title={Equivariant diffusion for molecule generation in 3d},
  author={Hoogeboom, Emiel and Satorras, V{\i}ctor Garcia and Vignac, Cl{\'e}ment and Welling, Max},
  booktitle={International conference on machine learning},
  pages={8867--8887},
  year={2022},
  organization={PMLR}
}

@inproceedings{nextmol,
	title     = {{NExT-Mol}: 3D Diffusion Meets 1D Language Modeling for 3D Molecule Generation},
	author    = {Liu, Zhiyuan and Luo, Yanchen and Huang, Han and Zhang, Enzhi and Li, Sihang and Fang, Junfeng and Shi, Yaorui and Wang, Xiang and Kawaguchi, Kenji and Chua, Tat-Seng},
	booktitle = {International Conference on Learning Representations (ICLR)},
	year      = {2025},
	url       = {https://openreview.net/forum?id=p66a00KLWN}
}

@article{propmolflow,
	title     = {{PropMolFlow}: property-guided molecule generation with geometry-complete flow matching},
	author    = {Zeng, Cheng and Jin, Jirui and Ambrose, Connor and Karypis, George and Transtrum, Mark K. and Tadmor, Ellad B. and Hennig, Richard G. and Roitberg, Adrian and Martiniani, Stefano and Liu, Mingjie},
	journal   = {Nature Computational Science},
	volume    = {6},
	number    = {3},
	pages     = {233--242},
	year      = {2026},
	doi       = {10.1038/s43588-025-00946-y}
}

@inproceedings{geoldm,
  title={Geometric latent diffusion models for 3d molecule generation},
  author={Xu, Minkai and Powers, Alexander S and Dror, Ron O and Ermon, Stefano and Leskovec, Jure},
  booktitle={International Conference on Machine Learning},
  pages={38592--38610},
  year={2023},
  organization={PMLR}
}

@article{CCMO,
	title={A coevolutionary framework for constrained multiobjective optimization problems},
	author={Tian, Ye and Zhang, Tao and Xiao, Jianhua and Zhang, Xingyi and Jin, Yaochu},
	journal={IEEE Transactions on Evolutionary Computation},
	volume={25},
	number={1},
	pages={102--116},
	year={2020},
	publisher={IEEE}
}

@article{cmoeacd,
	title={Constraint-Pareto Dominance and Diversity Enhancement Strategy based Evolutionary Algorithm for Solving Constrained Multiobjective Optimization Problems},
	author={Liu, Zhe and Han, Fei and Ling, Qinghua and Han, Henry and Jiang, Jing},
	journal={IEEE Transactions on Evolutionary Computation},
	year={2025},
	publisher={IEEE}
}

@inproceedings{geodiff,
	title     = {{GeoDiff}: A Geometric Diffusion Model for Molecular Conformation Generation},
	author    = {Xu, Minkai and Yu, Lantao and Song, Yang and Shi, Chence and Ermon, Stefano and Tang, Jian},
	booktitle = {International Conference on Learning Representations},
	year      = {2022},
}

@article{mdm,
	title   = {{MDM}: Molecular Diffusion Model for {3D} Molecule Generation},
	author  = {Huang, Lei and Zhang, Hengtong and Xu, Tingyang and Wong, Ka-Chun},
	journal = {Proceedings of the AAAI Conference on Artificial Intelligence},
	volume  = {37},
	number  = {4},
	pages   = {5105--5112},
	year    = {2023},
}

@inproceedings{moldiff,
	title     = {{MolDiff}: Addressing the Atom-Bond Inconsistency Problem in {3D} Molecule Diffusion Generation},
	author    = {Peng, Xingang and Guan, Jiaqi and Liu, Qiang and Ma, Jianzhu},
	booktitle = {Proceedings of the 40th International Conference on Machine Learning},
	series    = {Proceedings of Machine Learning Research},
	volume    = {202},
	pages     = {27611--27629},
	publisher = {PMLR},
	year      = {2023},
}

@article{murray2009rise,
	title={The rise of fragment-based drug discovery},
	author={Murray, Christopher W. and Rees, David C.},
	journal={Nature Chemistry},
	volume={1},
	pages={187--192},
	year={2009}
}

@article{imrie2020deep,
	title={Deep generative models for {3D} linker design},
	author={Imrie, Fergus and Bradley, Anthony R. and van der Schaar, Mihaela and Deane, Charlotte M.},
	journal={Journal of Chemical Information and Modeling},
	volume={60},
	number={4},
	pages={1983--1995},
	year={2020}
}

@inproceedings{rosenthal2020diversity,
	title     = {Diversity Promoting Strategies in a Multi- and Many-Objective Evolutionary Algorithm for Molecular Optimization},
	author    = {Rosenthal, Susanne},
	booktitle = {Bioinspired Optimization Methods and Their Applications},
	series    = {Lecture Notes in Computer Science},
	volume    = {12438},
	pages     = {294--307},
	year      = {2020},
	publisher = {Springer},
	address   = {Cham},
	doi       = {10.1007/978-3-030-63710-1_23}
}

@article{tanabe2020review,
	title={A Review of Evolutionary Multimodal Multiobjective Optimization},
	author={Tanabe, Ryoji and Ishibuchi, Hisao},
	journal={IEEE Transactions on Evolutionary Computation},
	volume={24},
	number={1},
	pages={193--200},
	year={2020},
	doi={10.1109/TEVC.2019.2909744}
}

@article{jensen2019graph,
	title   = {A Graph-Based Genetic Algorithm and Generative Model/Monte Carlo Tree Search for the Exploration of Chemical Space},
	author  = {Jensen, Jan H.},
	journal = {Chemical Science},
	volume  = {10},
	number  = {12},
	pages   = {3567--3572},
	year    = {2019},
	doi     = {10.1039/C8SC05372C}
}

@article{janus2022,
	title   = {Parallel Tempered Genetic Algorithm Guided by Deep Neural Networks for Inverse Molecular Design},
	author  = {Nigam, AkshatKumar and Pollice, Robert and Aspuru-Guzik, Al{\'a}n},
	journal = {Digital Discovery},
	volume  = {1},
	number  = {4},
	pages   = {390--404},
	year    = {2022},
	doi     = {10.1039/D2DD00003B}
}

@article{verhellen2022pareto,
	title   = {Graph-Based Molecular Pareto Optimisation},
	author  = {Verhellen, Jonas},
	journal = {Chemical Science},
	volume  = {13},
	number  = {25},
	pages   = {7526--7535},
	year    = {2022},
	doi     = {10.1039/D2SC00821A}
}

@article{momo2024,
	title   = {Evolutionary Multiobjective Molecule Optimization in an Implicit Chemical Space},
	author  = {Xia, Xin and Liu, Yiping and Zheng, Chunhou and Zhang, Xingyi and Wu, Qingwen and Gao, Xin and Zeng, Xiangxiang and Su, Yansen},
	journal = {Journal of Chemical Information and Modeling},
	volume  = {64},
	number  = {13},
	pages   = {5161--5174},
	year    = {2024},
	doi     = {10.1021/acs.jcim.4c00031}
}

@inproceedings{molleo2025,
	title     = {Efficient Evolutionary Search Over Chemical Space with Large Language Models},
	author    = {Wang, Haorui and Skreta, Marta and Ser, Cher-Tian and Gao, Wenhao and Kong, Lingkai and Strieth-Kalthoff, Felix and Duan, Chenru and Zhuang, Yuchen and Yu, Yue and Zhu, Yanqiao and Du, Yuanqi and Aspuru-Guzik, Al{\'a}n and Neklyudov, Kirill and Zhang, Chao},
	booktitle = {International Conference on Learning Representations},
	year      = {2025}
}

@article{autogrow4,
	title   = {{AutoGrow4}: An Open-Source Genetic Algorithm for De Novo Drug Design and Lead Optimization},
	author  = {Spiegel, Jacob O. and Durrant, Jacob D.},
	journal = {Journal of Cheminformatics},
	volume  = {12},
	number  = {1},
	pages   = {25},
	year    = {2020},
	doi     = {10.1186/s13321-020-00429-4}
}

@inproceedings{rga2022,
	title     = {Reinforced Genetic Algorithm for Structure-Based Drug Design},
	author    = {Fu, Tianfan and Gao, Wenhao and Coley, Connor W. and Sun, Jimeng},
	booktitle = {Advances in Neural Information Processing Systems},
	volume    = {35},
	pages     = {12325--12338},
	year      = {2022}
}

@inproceedings{mudiff,
	title     = {{MUDiff}: Unified Diffusion for Complete Molecule Generation},
	author    = {Hua, Chenqing and Luan, Sitao and Xu, Minkai and Ying, Zhitao and Fu, Jie and Ermon, Stefano and Precup, Doina},
	booktitle = {Proceedings of the Second Learning on Graphs Conference},
	series    = {Proceedings of Machine Learning Research},
	volume    = {231},
	pages     = {33:1--33:26},
	publisher = {PMLR},
	year      = {2024},
}

@inproceedings{targetdiff,
	title     = {{3D} Equivariant Diffusion for Target-Aware Molecule Generation and Affinity Prediction},
	author    = {Guan, Jiaqi and Qian, Wesley Wei and Peng, Xingang and Su, Yufeng and Peng, Jian and Ma, Jianzhu},
	booktitle = {International Conference on Learning Representations},
	year      = {2023},
}

@article{diffsbdd,
	title   = {Structure-Based Drug Design with Equivariant Diffusion Models},
	author  = {Schneuing, Arne and Harris, Charles and Du, Yuanqi and Didi, Kieran and Jamasb, Arian and Igashov, Ilia and Du, Weitao and Gomes, Carla and Blundell, Tom L. and Lio, Pietro and Welling, Max and Bronstein, Michael and Correia, Bruno},
	journal = {Nature Computational Science},
	volume  = {4},
	number  = {12},
	pages   = {899--909},
	year    = {2024},
	doi     = {10.1038/s43588-024-00737-x}
}

@inproceedings{decompdiff,
	title     = {{DecompDiff}: Diffusion Models with Decomposed Priors for Structure-Based Drug Design},
	author    = {Guan, Jiaqi and Zhou, Xiangxin and Yang, Yuwei and Bao, Yu and Peng, Jian and Ma, Jianzhu and Liu, Qiang and Wang, Liang and Gu, Quanquan},
	booktitle = {Proceedings of the 40th International Conference on Machine Learning},
	series    = {Proceedings of Machine Learning Research},
	volume    = {202},
	pages     = {11827--11846},
	publisher = {PMLR},
	year      = {2023},
}

@data{cd2020,
doi = {10.21227/45c9-vg74},
url = {https://dx.doi.org/10.21227/45c9-vg74},
author = {Yaoxiang Zhang},
publisher = {IEEE Dataport},
title = {CrossDocked2020},
year = {2024} }

@article{erlanson2016twenty,
  title={Twenty years on: the impact of fragments on drug discovery},
  author={Erlanson, Daniel A and Fesik, Stephen W and Hubbard, Roderick E and Jahnke, Wolfgang and Jhoti, Harren},
  journal={Nature Reviews Drug Discovery},
  volume={15},
  number={9},
  pages={605--619},
  year={2016},
  publisher={Nature Publishing Group UK London}
}

@article{bekes2022protac,
  title={{PROTAC} targeted protein degraders: the past is prologue},
  author={B{\'e}k{\'e}s, Mikl{\'o}s and Langley, David R and Crews, Craig M},
  journal={Nature Reviews Drug Discovery},
  volume={21},
  number={3},
  pages={181--200},
  year={2022},
  publisher={Nature Publishing Group UK London}
}

@article{posebuster,
	title = {{{PoseBusters}}: {{AI-based}} Docking Methods Fail to Generate Physically Valid Poses or Generalise to Novel Sequences},
	shorttitle = {{{PoseBusters}}},
	author = {Buttenschoen, Martin and Morris, Garrett M. and Deane, Charlotte M.},
	year = "2024",
	journal = "Chemical Science",
	volume = "15",
	issue = "9",
	pages = "3130-3139",
	publisher = "The Royal Society of Chemistry",
	doi = "10.1039/D3SC04185A",
	url = "http://dx.doi.org/10.1039/D3SC04185A",
}

@article{spea2,
  title={SPEA2: Improving the strength Pareto evolutionary algorithm},
  author={Zitzler, Eckart and Laumanns, Marco and Thiele, Lothar},
  journal={TIK report},
  volume={103},
  year={2001},
  publisher={ETH Zurich}
}

@article{drugs,
	title={GEOM, energy-annotated molecular conformations for property prediction and molecular generation},
	author={Axelrod, Simon and Gomez-Bombarelli, Rafael},
	journal={Scientific Data},
	volume={9},
	number={1},
	pages={185},
	year={2022},
	publisher={Nature Publishing Group UK London}
}

@article{tfg,
  title={Tfg: Unified training-free guidance for diffusion models},
  author={Ye, Haotian and Lin, Haowei and Han, Jiaqi and Xu, Minkai and Liu, Sheng and Liang, Yitao and Ma, Jianzhu and Zou, James Y and Ermon, Stefano},
  journal={Advances in Neural Information Processing Systems},
  volume={37},
  pages={22370--22417},
  year={2024}
}

@article{zx1,
  title={Multi-objective evolutionary computation for topology coverage assessment problem},
  author={Zhou, Xing and Wang, Huaimin and Ding, Bo and Peng, Wei and Wang, Rui},
  journal={Knowledge-Based Systems},
  volume={177},
  pages={1--10},
  year={2019},
  publisher={Elsevier}
}

@article{zx2,
  title={Solving multi-scenario cardinality constrained optimization problems via multi-objective evolutionary algorithms},
  author={Zhou, Xing and Wang, Huaimin and Peng, Wei and Ding, Bo and Wang, Rui},
  journal={Science China Information Sciences},
  volume={62},
  pages={1--18},
  year={2019},
  publisher={Springer}
}

@article{gcdm,
  title={Geometry-complete diffusion for 3D molecule generation and optimization},
  author={Morehead, Alex and Cheng, Jianlin},
  journal={Communications Chemistry},
  volume={7},
  number={1},
  pages={150},
  year={2024},
  publisher={Nature Publishing Group UK London}
}

@article{eegsde,
  title={Equivariant energy-guided sde for inverse molecular design},
  author={Bao, Fan and Zhao, Min and Hao, Zhongkai and Li, Peiyao and Li, Chongxuan and Zhu, Jun},
  journal={arXiv preprint arXiv:2209.15408},
  year={2022}
}

@article{qm9,
  title={Quantum chemistry structures and properties of 134 kilo molecules},
  author={Ramakrishnan, Raghunathan and Dral, Pavlo O and Rupp, Matthias and Von Lilienfeld, O Anatole},
  journal={Scientific data},
  volume={1},
  number={1},
  pages={1--7},
  year={2014},
  publisher={Nature Publishing Group}
}


\end{document}